\documentclass[12pt]{article}

\usepackage{amsmath,amssymb,amsfonts,amsthm}
\usepackage{graphicx}
\usepackage{geometry}
\usepackage{float}
\usepackage{caption}
\usepackage{subcaption}
\usepackage{multirow}
\usepackage{hyperref}
\usepackage{cite}

\usepackage[utf8]{inputenc}
\usepackage[T1]{fontenc}
\usepackage{amsmath, amsfonts, amssymb, amsthm}
\usepackage{geometry}
\newtheorem{theorem}{Theorem}[section]
\newtheorem{lemma}[theorem]{Lemma}
\newtheorem{definition}[theorem]{Definition}

\newtheorem{assumption}[theorem]{Assumption}
\newtheorem{claim}[theorem]{Claim}
\begin{document}

\title{New Fourth-Order Grayscale Indicator-Based Telegraph Diffusion Model for Image Despeckling: Theory and Applications }

\author{
    \textbf{Rajendra K. Ray\textsuperscript{1}\thanks{Corresponding author: \texttt{rajendra@iitmandi.ac.in}}, Manish Kumar\textsuperscript{1}, and Ananta K. Majee\textsuperscript{2}} \\[2ex]
    \textsuperscript {1}School of Mathematical and Statistical Sciences, \\
    Indian Institute of Technology Mandi, Mandi (H.P), 175075, India \\[1ex]
    \textsuperscript{2}Department of Mathematics, \\
    Indian Institute of Technology Delhi, Academic Complex West (99B), \\
    Hauz Khas, New Delhi-110016, India \\[2ex]
\textsuperscript{1}\texttt{rajendra@iitmandi.ac.in}, \textsuperscript{1}\texttt{manishkumarkitlana@gmail.com}, \\
    \textsuperscript{2}\texttt{majee@maths.iitd.ac.in}
}

\date{}

\maketitle

\begin{abstract}

Second-order models have been widely used for suppressing multiplicative noise, but they often introduce blocky artifacts in the early stages of denoising. To resolve this, we propose a fourth-order nonlinear PDE model that integrates diffusion and wave properties. The diffusion process, guided by both the Laplacian and intensity values, reduces noise better than gradient-based methods, while the wave nature handle the oscillation and fine details and textures. Moreover, We establish the existence of a weak solution to the proposed model by applying Schauder’s fixed point theorem. The effectiveness of the proposed model is evaluated against two second-order anisotropic diffusion approaches using Peak Signal-to-Noise Ratio (PSNR) and Mean Structural Similarity Index (MSSIM) for images with available ground truth. For SAR images, the Speckle Index (SI) is used to measure noise reduction. Additionally, we extend the proposed model to study color images, preserving both structure and color consistency. The same quantitative metrics PSNR and MSSIM are used for performance evaluation, ensuring a fair comparison across grayscale and color images. In all the cases, our computed results produce better results compared to existing models in this genre.

\end{abstract}

\section{Introduction}

Image restoration plays a crucial role in image processing by reconstructing original images from noisy observations. In real-world applications, noise can degrade images in additive, multiplicative, or mixed forms, making denoising an essential preprocessing step for tasks like segmentation and pattern recognition. This study focuses on multiplicative speckle noise, a common issue in synthetic aperture radar (SAR) and ultrasound imaging. The multiplicative noise is mathematically represented as:\begin{equation}N = I\eta,\end{equation}where $I$ is the original image, $\eta$ denotes noise, and $N$ is the noisy image observed. In the case of multiplicative speckle noise, $\eta$ follows a gamma distribution with mean 1 and variance $1/L$, where $L$ represents the number of observations:\begin{equation}p(\eta) =\begin{cases}\frac{1}{\Gamma(L)} L^L \eta^{L-1} e^{-L\eta}, & \eta > 0, \\0, & \eta = 0.\end{cases}\end{equation}Unlike additive noise, multiplicative noise varies with intensity, making brighter regions more vulnerable to degradation. This complexity makes speckle noise reduction particularly challenging. Over the years, variational and PDE-based methods have been extensively explored to address image denoising \cite{shi2008nonlinear,aubert2008variational,zhou2014degenerate,siddig2018fourth}. The AA model by Aubert and Aujol \cite{aubert2008variational} introduced a regularization-based approach, but its non-convex nature posed computational difficulties. To resolve this, Shi and Osher \cite{shi2008nonlinear} transformed the multiplicative noise into an additive form using logarithmic functions, leading to a globally convex model. Further advancements, such as the novel total variation model (NTV) \cite{huang2009total}, improved edge preservation and denoising efficiency. In PDE-based methods, Zhou et al. \cite{zhou2014degenerate} introduced a Doubly Degenerate (DD) diffusion model, which used gray-level indicators in the diffusion coefficient but lacked well-posedness. Later, Shan et al. \cite{shan2019smooth} refined this model to ensure stability. Recent approaches have combined diffusion and wave dynamics to better preserve high-frequency details while reducing noise. However, second-order PDEs tend to produce blocky artifacts in denoised images. Majee et al. \cite{Majee} have proposed effective methods for image denoising, addressing multiplicative speckle noise. Their approaches demonstrate significant improvements in noise reduction while preserving essential image details. Fourth-order partial differential equation (PDE) models have been widely investigated for image restoration and denoising tasks. An early contribution in this direction was presented in \cite{you2000fourth}, where a fourth-order regularization framework was introduced for noise removal. Subsequently, related approaches for image restoration based on fourth-order PDEs have been explored in \cite{Ma2007}. Variational formulations of fourth-order PDE models have also been developed for effective image denoising, as discussed in \cite{Kim2009}. Further improvements in diffusion-based fourth-order models were proposed in \cite{Guidotti2011} to enhance denoising performance. In addition, adaptive frameworks that integrate both second-order and fourth-order terms have been studied in \cite{Adaptive2017}. A fourth-order telegraph–diffusion framework for multiplicative noise removal is also reported in \cite{Padmaja2025}. More recently, Roy et al. \cite{roy2024local} developed a Bayesian MAP framework integrating total variation regularization to effectively model speckle noise while preserving edges. Li et al. \cite{li2022novel} proposed a hybrid PDE model, blending total variation filtering with anisotropic diffusion, achieving better despeckling performance in SAR images. Cuomo et al. \cite{cuomo2024learned} introduced a learned variational model leveraging deep learning to suppress adaptive noise in ultrasound and satellite imaging.
Perona and Malik \cite{perona1990scale} introduced the first anisotropic diffusion model in 1990 to address Gaussian additive noise. The governing equation of the Perona-Malik model is given by:\begin{equation}\frac{\partial I}{\partial t} = \text{div} \left( C(|\nabla I|) \nabla I \right),\end{equation}
where $I$ represents the image, $t$ denotes time, and $C(|\nabla I|)$ is the diffusion coefficient that depends on the gradient magnitude. This function, which is smooth and non-increasing, satisfies $C(0) = 1$ and $\lim_{x \to \infty} C(x) = 0$. The model effectively preserves edges but may retain or amplify high-frequency noise, limiting its denoising capabilities. To enhance its performance, several extensions have been proposed. Zhou et al. \cite{zhou2014degenerate} introduced a Doubly Degenerate (DD) model that incorporates both grayscale indicators and gradients, improving image denoising compared to purely gradient-based models. The governing equation is:\begin{equation}\frac{\partial I}{\partial t} = \text{div} \left( C(I, |\nabla I|) \nabla I \right), \quad \Omega_T = \Omega \times (0, T),\end{equation}with the boundary and initial conditions:\begin{equation}\frac{\partial I}{\partial n} = 0, \quad \text{on } \partial \Omega_T = \partial \Omega \times (0, T),\end{equation}\begin{equation}I(a, 0) = f(a), \quad \text{in } \Omega.\end{equation} The diffusion coefficient is given by:\begin{equation}C(I, |\nabla I|) = \frac{2 |I|^\alpha}{M^\alpha + |I|^\alpha} \cdot \frac{1}{(1 + |\nabla I|^2)^{(1 - \beta)/2}},\end{equation}where $\beta$ regulates gradient sensitivity. Shan et al. \cite{shan2019smooth} introduced a regularized version of the DD model to ensure well-posedness:\begin{equation}\frac{\partial I}{\partial t} = \nabla \cdot \left( C(I_\xi, |\nabla I_\xi|)\nabla I \right),\end{equation}with boundary and initial conditions:\begin{equation}\frac{\partial I}{\partial n} = 0, \quad \text{on } \partial \Omega_T,\end{equation}\begin{equation}I(a, 0) = f(a), \quad \text{in } \Omega.\end{equation}where $I_\xi$ is a Gaussian-smoothed image, and the diffusion coefficient is defined as:\begin{equation}C(I_\xi, |\nabla I_\xi|) = \left(\frac{I_\xi}{M_\xi }\right)^\alpha \cdot \frac{1}{1 + |\nabla I_\xi|^\nu}.\end{equation}To further enhance despeckling performance, Majee et al. \cite{Majee} proposed the TDE model, which combines the diffusion and wave equations. The governing equation for the TDE model is:\begin{equation}\frac{\partial^2 I}{\partial t^2} + \gamma \frac{\partial I}{\partial t} = \nabla \cdot \left( C(I_\xi, |\nabla I_\xi|) \nabla I \right),\end{equation}where $\gamma$ regulates wave behavior. The initial and boundary conditions are:\begin{equation}I(a, 0) = f(a), \quad \frac{\partial I}{\partial t}(a, 0) = 0, \quad \text{in } \Omega,\end{equation}\begin{equation}\frac{\partial I}{\partial n} = 0, \quad \text{on } \partial \Omega_T.\end{equation}The diffusion coefficient in the TDE model is expressed as:\begin{equation}C(I_\xi, |\nabla I_\xi|) = \frac{2 |I_\xi|^\alpha}{M_\xi^\alpha + |I_\xi|^\alpha} \cdot \frac{1}{1 + (|\nabla I_\xi| / k)^2},\end{equation}where $M_\xi = \max_{x \in \Omega} | I_\xi(x,t) |$, $\alpha$ is a parameter that regulates intensity adaptation, and $k$ controls the gradient-based diffusion. The inclusion of wave dynamics in the TDE model helps in maintaining fine details and providing improved image restoration results. Higher-order PDEs have been explored to overcome the limitations of second-order models. You and Kaveh \cite{you2000fourth} replaced the gradient with the Laplacian, resulting in the fourth-order equation:\begin{equation}\frac{\partial I}{\partial t} = -\Delta \left( C(\Delta I) \Delta I \right),\end{equation}with boundary and initial conditions:\begin{equation}\frac{\partial I}{\partial n} = 0,\frac{\partial (\Delta I)}{\partial n} = 0 \quad \text{on } \partial \Omega_T,\end{equation}\begin{equation}I(a, 0) = f(a), \quad \text{in } \Omega.\end{equation}where the diffusion coefficient is:\begin{equation}C(|\Delta I|) = \frac{1}{1 + (|\Delta I|/k)^2}.\end{equation}This approach improves noise reduction while preserving edges. We propose a novel fourth-order nonlinear PDE model for multiplicative speckle noise, achieving superior results in real SAR images.

\section{Proposed Model}\label{sec:proposed_model}
Inspired by\cite{you2000fourth, Majee} we developed a nonlinear fourth-order model that effectively integrates wave dynamics and diffusion to balance noise reduction with structure preservation. The governing equation is:\begin{equation}\label{eq:main_pde}
\frac{\partial^2 I}{\partial t^2} + \gamma \frac{\partial I}{\partial t} = -\Delta \left( C(I_\xi, |\Delta I_\xi|) \Delta I\right) - \lambda  \left( \frac{I - f}{I} \right)^2, \quad \forall (x, t) \in \Omega \times (0, T),\end{equation}with initial and boundary conditions:\begin{equation}I(a, 0) = f(a), \quad I_t(a, 0) = 0, \quad \forall x \in \Omega,
\end{equation}
\begin{equation}\label{eq:bc_ic}
\frac{\partial I}{\partial n} = 0, \quad \frac{\partial (\Delta I)}{\partial n} = 0, \quad \forall (x,t) \in \partial \Omega \times (0,T).
\end{equation}
Here, $f$ is the noisy image, and $t$ represents time.
The diffusion coefficient is defined as:
\begin{equation}C(I_\sigma, |\Delta I_\sigma|) = \frac{2 |I_\xi|^\nu}{M_\xi^\nu+ |I_\sigma|^\nu} \cdot \frac{1} {1 + (|\Delta I_\sigma|/k)^2},\end{equation}

where $M_\sigma = \max_{x \in \Omega} $$| I_\sigma(x,t) |$ represents the maximum intensity, and $\alpha$ and $k$ are parameters. The term $\frac{2 |I_\sigma|^\nu}{M_\sigma^\alpha + |I_\sigma|^\nu}$ adapts diffusion to intensity variations, reducing smoothing in regions with significant features to preserve edges. The factor $\frac{1}{1 + (|\Delta I_\sigma|/k)^2}$ further modulates diffusion based on the Laplacian, slowing diffusion in high-gradient regions. Gaussian smoothing, $I_\sigma = G_\sigma \ast I$, stabilizes the model by reducing noise while retaining fine details.
\section{Well-posedness of the weak solution}\label{sec:well_posedness}
In this section, we prove the existence and uniqueness of the weak solution of the proposed model \eqref{eq:main_pde}--\eqref{eq:bc_ic}. Since the problem \eqref{eq:main_pde}--\eqref{eq:bc_ic} is nonlinear, we first consider the linearized problem and then use the Schauder fixed-point theorem to show the existence of a weak solution.
\subsection{Technical framework and statement of the main result}
Throughout this section, $C$ denotes a generic positive constant. For $1 \le p \le \infty$, we denote by $(L^p(\Omega), \|\cdot\|_{L^p(\Omega)})$ the standard spaces of $p$-th order Lebesgue integrable functions on $\Omega$. For $r \in \mathbb{N}$, we write $(H^r(\Omega), \|\cdot\|_{H^r(\Omega)})$ for the usual Sobolev Hilbert spaces on $\Omega$, and $(H^r(\Omega))'$ for the dual space of $H^r(\Omega)$. For any Banach space ${\tt X}$, the spaces $L^p(0,T; {\tt X})$ denote the standard Bochner spaces for any $p\ge 1$. Throughout this article, to simplify the notation, we write $L^p$, $H^r$, and $(H^r)'$ instead of $L^p(\Omega)$, $H^r(\Omega)$, and $(H^r(\Omega))'$, respectively. 
\vspace{0.1cm}

We introduce the solution space $\mathcal{W}(0, T)$ for the weak solution of the proposed model \eqref{eq:main_pde}--\eqref{eq:bc_ic}, defined as:
\begin{equation*}
    \mathcal{W}(0, T) = \left\{ I \in L^\infty(0, T; H^2) \;\big|\; I_t \in L^\infty(0, T; L^2), \; I_{tt} \in L^2(0, T; (H^2)') \right\}.
\end{equation*}
Note that the space $\mathcal{W}(0, T)$, equipped with its natural norm, is a Banach space.

\begin{definition}[Weak Solution] \label{def:weak_solution}
A function $I$ is called a weak solution of the problem \eqref{eq:main_pde}-\eqref{eq:bc_ic} if:
\begin{itemize}
    \item[(a)] $I \in W(0,T)$ and satisfies the initial conditions \eqref{eq:bc_ic} .
    \item[(b)] For all test functions $\phi \in H^2(\Omega)$ and a.e. $t \in (0,T)$, there holds:
\begin{equation} \label{eq:weak_form}
 \langle I_{tt}, \phi \rangle
 + \gamma (I_t, \phi)_{L^2}
 + \int_{\Omega} C(I_\sigma, |\Delta I_\sigma|) \Delta I \Delta \phi \, dx
 + \left( \lambda \left(\frac{I-f}{I}\right)^2, \phi \right)_{L^2} = 0.
 \end{equation}

\end{itemize}
\end{definition}
Our aim is to establish the well-posedness of the weak solutions, and we will do so under the following assumption:

\begin{assumption}[\textbf{A.1}] \label{ass:A1}
The initial data $f \in H^2(\Omega)$ is strictly bounded below away from zero i.e., there exists $\alpha>0$ such that
\[
0 < \alpha := \inf_{x \in \Omega} f(x).
\]
\end{assumption}

\begin{theorem} \label{thm:main_existence}
Let Assumption \ref{ass:A1} be true. Then the problem \eqref{eq:main_pde}-\eqref{eq:bc_ic} admits a unique weak solution in the sense of Definition \ref{def:weak_solution}.
\end{theorem}

\subsection{Linearized problem and existence of a weak solution}
As mentioned earlier, due to the nonlinear nature of the problem, we first consider linear auxiliary problem and study its wellposedness result. To proceed, we first consider the following space: 
for any $M > 0$, define the convex set of admissible functions 
\begin{equation} \label{eq:set_WM}
W_{M} = \left\{ \bar{I} \in W(0,T) \; : \; 
\begin{aligned}
& \|\bar{I}\|_{L^\infty(0,T; H^2)} + \|\bar{I}_t\|_{L^\infty(0,T; L^2)} \le M \|f\|_{H^2}, \\
& 0 < \alpha \le \bar{I}(x,t) \text{ for a.e. } (x,t) \in \Omega_T
\end{aligned}
\right\}.
\end{equation}

For any $\bar{I} \in W_{M}$, consider the \textbf{linearized problem}:
\begin{equation} \label{eq:linearized_pde}
I_{tt} + \gamma I_t + \Delta (\bar{g}(x,t) \Delta I) + \mathcal{S}(x,t) = 0 \quad \text{in } \Omega_T,
\end{equation}
satisfying the initial conditions \eqref{eq:bc_ic}, where the frozen coefficient function $\bar{g}$ and source term $\mathcal{S}$ are given by:
\begin{equation} \label{eq:frozen_coeffs}
\bar{g}(x,t):= C(\bar{I}_\sigma(x,t), |\Delta \bar{I}_\sigma(x,t)|) \quad \text{and} \quad \mathcal{S}(x,t):= \lambda \left(\frac{\bar{I}(x,t)-f(x)}{\bar{I}(x,t)}\right)^2.
\end{equation}
\begin{claim} \label{claim:bounds}
There exist positive constants $\kappa, C_{coef}, C_{\mathcal{S}} > 0$ depending only on the initial data $f$, parameters $M, k,G_\sigma, \alpha$, and the domain $\Omega$ such that for any $\overline{I} \in W_M$:
\begin{enumerate}
    \item[{\rm (i)}] $0 < \kappa \le \bar{g}(x,t) \le 2$,
    \item[{\rm (ii)}] $|\bar{g}_t(x,t)| \le C_{coef}$,
    \item[{\rm (iii)}] $\|\mathcal{S}(x,t)\|_{L^2(\Omega_T)} \le C_{\mathcal{S}}$.
\end{enumerate}
\end{claim}

\begin{proof}

The proofs establishing the upper and lower bounds in (i), as well as the bound on the time derivative in (ii), follow exactly the arguments presented for Claim 3.1 in \cite{Majee}. We thus omit the detailed derivation here and refer the reader to \cite{Majee} for the complete proof.

(iii) Since $\bar{I} \in W_M$, we have the bound $\bar{I}(x,t) \ge \alpha > 0$. Thus:
\[
|\mathcal{S}(x,t)| = \lambda \left| 1 - \frac{f(x)}{\bar{I}(x,t)} \right|^2 \le \lambda \left( 1 + \frac{|f(x)|}{\alpha} \right)^2 \le \lambda\left( 1 + \frac{\|f\|_{H^2}}{\alpha}\right)\,.
\]
Therefore, there exists a constant $C_{\mathcal{S}}>0$ such that
\[
\|\mathcal{S}\|_{L^2(\Omega_T)} \le C_{\mathcal{S}}.
\]
This finishes the proof of the claim.
\end{proof}

\begin{lemma}\label{lemma:linear_combined}
For any fixed $\bar{I} \in W_M$, the linearized problem 
\begin{equation} \label{eq:linearized_pde2}
    I_{tt} + \gamma I_t + \Delta(\bar{g}(x,t)\Delta I) + \mathcal{S}(x,t) = 0 \quad \text{in } \Omega_T,
\end{equation}
subject to the initial and boundary conditions, admits a unique weak solution $I \in W(0,T)$. Furthermore, this solution satisfies the following energy estimates: for a constant $C > 0$ depending only on the parameters $M, \alpha, \gamma, k$, and the initial data $f$ such that
\begin{enumerate}
    \item[(a)] $\|I\|_{L^\infty(0,T; H^2(\Omega))} + \|I_t\|_{L^\infty(0,T; L^2(\Omega))} \le C \|f\|_{H^2(\Omega)}$.
    \item[(b)] $\|I_{tt}\|_{L^2(0,T; (H^2(\Omega))')} \le C \|f\|_{H^2(\Omega)}$.
\end{enumerate}
\end{lemma}

\begin{proof}
Thanks to Claim 2.4, one can apply the classical Galerkin method to show that there exists a unique weak solution $I \in \mathcal{W}(0, T)$ of the linearized problem \eqref{eq:linearized_pde} with the initial condition \eqref{eq:bc_ic}.
Next we prove that the solution of the linearized problem enjoys certain regularity:
 \paragraph{Proof of (a).} We begin by multiplying the linearized equation \eqref{eq:linearized_pde} by $I_t(x,t)$, and integrating by parts to have
\begin{equation} \label{eq:B.1}
\frac{1}{2} \frac{d}{dt} \|I_t(t)\|_{L^2}^2 + \gamma \|I_t(t)\|_{L^2}^2 + \int_{\Omega} \Delta (\bar{g} \Delta I) I_t \, dx = - \int_{\Omega} \mathcal{S} I_t \, dx.
\end{equation}
Thanks to Green's identity and the Neumann boundary conditions, we can re-write the third term of the l.h.s of \eqref{eq:B.1} as
\begin{equation}  \label{eq:B.2}
\int_{\Omega} \Delta (\bar{g} \Delta I) I_t \, dx = \int_{\Omega} \bar{g} \Delta I \Delta I_t \, dx = \frac{1}{2} \frac{d}{dt} \int_{\Omega} \bar{g} (\Delta I)^2 dx - \frac{1}{2} \int_{\Omega} \bar{g}_t (\Delta I)^2 dx.
\end{equation}

With the bounds established in Claim \ref{claim:bounds}, we observe that
\begin{equation} \label{eq:B.3}
\frac{1}{2} \left| \int_{\Omega} \bar{g}_t (\Delta I)^2 dx \right| \le \frac{C_{coef}}{2} \|\Delta I\|_{L^2}^2 \le \frac{C_{coef}}{2\kappa} \int_{\Omega} \bar{g} (\Delta I)^2 dx.
\end{equation}

For the source term, using the Cauchy-Schwarz and Young's inequalities:
\begin{equation}  \label{eq:B.4}
\left| \int_{\Omega} \mathcal{S} I_t \, dx \right| \le \frac{1}{2} \|\mathcal{S}\|_{L^2}^2 + \frac{1}{2} \|I_t\|_{L^2}^2.
\end{equation}

Now using \eqref{eq:B.2},\eqref{eq:B.3}, and \eqref{eq:B.4} in \eqref{eq:B.1}, and dropping the non-negative term $\gamma \|I_t\|_{L^2}^2$, we have
\begin{equation}  \label{eq:B.5}
\frac{d}{dt} \left[ \|I_t\|_{L^2}^2 + \int_{\Omega} \bar{g} (\Delta I)^2 dx \right] \le C \left( \|I_t\|_{L^2}^2 + \int_{\Omega} \bar{g} (\Delta I)^2 dx \right) + \|\mathcal{S}\|_{L^2}^2.
\end{equation}

The application of Gronwall's lemma gives the following: for a.e. $t \in (0, T]$,
\begin{equation} \label{eq:B.6}
\|I_t(t)\|_{L^2}^2 + \|\Delta I(t)\|_{L^2}^2 \le C \left( \|f\|_{H^2}^2 + \int_0^T \|\mathcal{S}(s)\|_{L^2}^2 ds \right) \le C.
\end{equation}
Since $I(x, t) = f(x) + \int_0^t I_t(s) ds$, we have by way of Young's inequality and \eqref{eq:B.6}
\begin{equation} \label{eq:B.7}
\|I(t)\|_{L^2}^2 \le 2 \|f\|_{L^2}^2 + 2T \int_0^t \big\| I_t(s) \big\|_{L^2}^2 ds \le C.
\end{equation}
By elliptic regularity for the Laplacian with Neumann boundary conditions, we know $\|I\|_{H^2} \le C_\Omega (\|I\|_{L^2} + \|\Delta I\|_{L^2})$. We combine this with \eqref{eq:B.6} and \eqref{eq:B.7} to conclude
\begin{equation}  \label{eq:B.8}
\|I\|_{L^\infty(0,T; H^2)} + \|I_t\|_{L^\infty(0,T; L^2)} \le C \|f\|_{H^2}.
\end{equation}
Hence (a) of the lemma follows.
\end{proof}
\vspace{0.5cm}

\paragraph{Proof of (b).} Multiplying the linearized $I$-equation by a test function $\phi \in H^2$ with $\|\phi\|_{H^2} \le 1$ and integrating over $\Omega$, we have
\begin{equation} \label{eq:weak_form_itt}
    \langle I_{tt}, \phi \rangle_{(H^2)', H^2} + \int_{\Omega} \left( \gamma I_t \phi + \bar{g} \Delta I \Delta \phi + \mathcal{S} \phi \right) dx = 0.
\end{equation}

We use the Cauchy-Schwarz inequality, along with the energy estimates from part (a) and the boundedness of $\bar{D}$, to obtain
\begin{align*}
    |\langle I_{tt}, \phi \rangle| &\le \left( \gamma \|I_t\|_{L^2} + \|\bar{g}\|_{L^\infty} \|\Delta I\|_{L^2} + \|\mathcal{S}\|_{L^2} \right) \|\phi\|_{H^2} \\
    &\le C \left( \|f\|_{H^2} + \|\mathcal{S}(t)\|_{L^2} \right) \|\phi\|_{H^2}.
\end{align*}

Hence, by the definition of the norm in the dual space $(H^2)'$, we get
\begin{equation} \label{eq:dual_norm_pointwise}
    \|I_{tt}(t)\|_{(H^2)'} \le C \left( \|f\|_{H^2} + \|\mathcal{S}(t)\|_{L^2} \right).
\end{equation}

Moreover, squaring both sides of \eqref{eq:dual_norm_pointwise} and integrating over $(0, T)$, and using the fact that the source term $\mathcal{S} \in L^2(0,T; L^2(\Omega))$, one arrives at the assertion that $\|I_{tt}\|_{L^2(0,T; (H^2)')} \le C\left( \|f\|_{H^2}\right)$. This completes the proof.

\begin{lemma} \label{lem:lower_bound}
Let $I \in W(0,T)$ be the unique weak solution of the linearized problem corresponding to $\overline{I} \in W_M$. If the initial data satisfies $f(x) \ge \alpha > 0$ for all $x \in \Omega$, then the solution of the linearized problem also satisfies:
\[
I(x,t) \ge \alpha \quad \text{for a.e. } (x,t) \in \Omega_T.
\]
\end{lemma}

\begin{proof}
To prove the assertion, we define
$$\theta(x,t) = (I(x,t) - \alpha)^- = \min(0, I(x,t) - \alpha).$$ Note that $\theta \le 0$ and $\theta \in L^2(0,T; H^2(\Omega))$.
Integrating the linearized PDE with respect to time and testing with $\theta$, we obtain:
\begin{equation} \label{eq:lower_bound_test}
\int_{\Omega} I_t \theta \, dx + \gamma \int_{\Omega} (I - f) \theta \, dx + \int_{\Omega} \left( \int_0^t \overline{g} \Delta I ds \right) \Delta \theta \, dx + \int_{\Omega} \left( \int_0^t \mathcal{S} ds \right) \theta \, dx = 0.
\end{equation}
For the time derivative term, we have
\[
\int_{\Omega} I_t \theta \, dx
= \frac{1}{2} \frac{d}{dt} \int_{\Omega} \theta^2 \, dx
= \frac{1}{2} \frac{d}{dt} \|\theta\|_{L^2(\Omega)}^2.
\]

For the damping term, since $f \ge \alpha$ and $I = \theta + \alpha$ on the support of $\theta$, we have
\[
(I - f)\theta
= (\theta + \alpha - f)\theta
= \theta^2 + (\alpha - f)\theta.
\]
Because $(\alpha - f) \le 0$ and $\theta \le 0$, the product $(\alpha - f)\theta$ is non-negative. Hence,
\[
\gamma \int_{\Omega} (I-f)\theta \, dx
\ge
\gamma \int_{\Omega} \theta^2 \, dx
=
\gamma \|\theta\|_{L^2(\Omega)}^2.
\]

For the diffusion term, by definition $\Delta \theta = \Delta I$ on the set where $I < \alpha$ and $\Delta \theta = 0$ otherwise. Since $\overline{g} \ge \kappa > 0$, we have
\[
\int_0^t \int_{\Omega} \overline{g} (\Delta \theta)^2 \, dx \, ds \ge 0.
\]

For the source term, since $\mathcal{S} \ge 0$ and $\theta \le 0$, we obtain
\[
\int_{\Omega} \left( \int_0^t \mathcal{S}(x,s) \, ds \right) \theta(x,t) \, dx \le 0.
\]
Moving this term to the right-hand side makes it non-negative. Discarding the non-negative diffusion and source contributions, we arrive at
\[
\frac{1}{2} \frac{d}{dt} \int_{\Omega} \theta^2 \, dx
+
\gamma \int_{\Omega} \theta^2 \, dx
\le 0.
\]
Because $I(x,0) = f(x) \ge \alpha$, we have
\[
\theta(x,0) = (f(x)-\alpha)^- = 0.
\]
By Gronwall's inequality,
\[
\|\theta(t)\|_{L^2(\Omega)}^2 = 0
\quad \text{for all } t \in [0,T].
\]
Thus $\theta = 0$ almost everywhere in $\Omega \times (0,T)$, and consequently
\[
I(x,t) \ge \alpha.
\]
\end{proof}
Before proving the existence of a weak solution, we need a technical lemma which we state in the following lemma. 
\begin{lemma} \label{lemma:coeff_prop}
Let $I_1, I_2 \in L^2(\Omega)$. For the diffusion coefficient, the following properties hold:
\begin{enumerate}
    \item[(i)] There exists a constant $C_0 > 0$, depending only on $\sigma$ and $\Omega$, such that:
    \begin{equation} \label{eq:bound_D}
    \|\Delta (G_\sigma \ast I_1)\|_{L^\infty(\Omega)} \le C_0 \|I_1\|_{L^2(\Omega)}.
    \end{equation}
    \item[(ii)] The mapping $I \mapsto C(I_\sigma, |\Delta I_\sigma|)$ is Lipschitz continuous i.e., there exists a constant $C > 0$, depending only on $\alpha, \sigma, \nu, K$, and $f$, such that for all $t\in [0,T]$
    \begin{equation} \label{eq:lipschitz_D}
    \|C(I_1) - C(I_2)\|_{L^\infty(\Omega)} \le C(\alpha, \xi, \nu, K, f) \|I_1 - I_2\|_{L^2(\Omega)},
    \end{equation}
    where $C(I_i):= C(I_{i,\sigma}, |\Delta I_{i,\sigma}|)$ with $I_{i, \sigma}:=G_\sigma * I_i~~(i=1,2)$.
\end{enumerate}
\end{lemma}

\begin{proof}
To prove (i), we utilize the properties of the Gaussian convolution and Young's inequality for convolutions with $p=2$ and $q=2$:
\begin{equation*}
\|\Delta (G_\sigma \ast I_1)\|_{L^\infty(\Omega)} \le \|\Delta G_\sigma\|_{L^2(\Omega)} \|I_1\|_{L^2(\Omega)} \le C_0 \|I_1\|_{L^2(\Omega)}\,.
\end{equation*}
Setting $C_0 = \|\Delta G_\sigma\|_{L^2(\Omega)}$, which depends only on $\sigma$ and the domain, we obtain the bound.
\paragraph{Proof of (ii).} We decompose the difference $g_1 - g_2$ into two terms:
\begin{align}
C(I_1) - C(I_2) &= \frac{|G_\sigma \ast I_1|^\nu}{(M_{I_1}^\sigma)^\nu + |G_\sigma \ast I_1|^\nu} 
\left( \frac{1}{1 + \left( \frac{|\Delta (G_\sigma \ast I_1)|}{K} \right)^2} - \frac{1}{1 + \left( \frac{|\Delta (G_\sigma \ast I_2)|}{K} \right)^2} \right) \nonumber \\
&\quad + \frac{1}{1 + \left( \frac{|\Delta (G_\sigma \ast I_2)|}{K} \right)^2} 
\left( \frac{|G_\sigma \ast I_1|^\nu}{(M_{I_1}^\sigma)^\nu + |G_\sigma \ast I_1|^\nu} - \frac{|G_\sigma \ast I_2|^\nu}{(M_{I_2}^\sigma)^\nu + |G_\sigma \ast I_2|^\nu} \right). \label{eq:diff_decomp}
\end{align}
By way of the convolution properties, we have
\begin{align}
    \| C(I_1) - C(I_2) \|_{L^\infty} &\le C(\xi, \nu, K,f) \|I_1 - I_2\|_{L^2} \nonumber \\
    &\quad + \left\| \frac{|G_\xi \ast I_1|^\nu (M_{I_2}^\xi)^\nu - |G_\xi \ast I_2|^\nu (M_{I_1}^\xi)^\nu}{\bigl( (M_{I_1}^\xi)^\nu + |G_\xi \ast I_1|^\nu \bigr) \bigl( (M_{I_2}^\xi)^\nu + |G_\xi \ast I_2|^\nu \bigr)} \right\|_{L^\infty} \nonumber \\
    &\equiv C(\xi, \nu, K, f) \|I_1 - I_2\|_{L^2} + \mathcal{A}. \label{eq:g_diff}
\end{align}
Since the initial data $f \ge \alpha > 0$ implies the uniform lower bound $(M_{I_i}^\xi)^\nu \ge (\alpha \|G_\xi\|_{L^1})^\nu$ for $i=1, 2$, we can bound the term $\mathcal{A}$ as follows:
\begin{align*}
    \mathcal{A} &\le \frac{1}{(\alpha \|G_\xi\|_{L^1})^{2\nu}} \left\| |G_\xi \ast I_1|^\nu (M_{I_2}^\xi)^\nu - |G_\xi \ast I_2|^\nu (M_{I_1}^\xi)^\nu \right\|_{L^\infty} \\
    &\le C(\alpha, \xi, \nu) \left\{ \left\| |G_\xi \ast I_1|^\nu \left( (M_{I_2}^\xi)^\nu - (M_{I_1}^\xi)^\nu \right) \right\|_{L^\infty} + \left\| (M_{I_1}^\xi)^\nu \left( |G_\xi \ast I_1|^\nu - |G_\xi \ast I_2|^\nu \right) \right\|_{L^\infty} \right\} \\
    &\le C(\alpha, \xi, \nu, f) \|I_1 - I_2\|_{L^2}.
\end{align*}
Combining the above relation with \eqref{eq:g_diff}, we obtain the required result. 
\end{proof}

\section{Proof of Theorem \ref{thm:main_existence}} In this section, we prove the well-posedness of the weak solution via the Generalized Schauder fixed-point theorem.
\subsection*{ Existence of weak solution} We introduce the convex subspace $W_0$ of $W(0, T)$, defined by
\begin{equation}
W_0 = \left\{ w \in W(0, T) : 
\begin{aligned}
& \|w\|_{L^\infty(0,T; H^2)} + \|w_t\|_{L^\infty(0,T; L^2)} \le M \|f\|_{H^2}; \\
& \|w_{tt}\|_{L^2(0,T; (H^2)')} \le C_{tt}; \\
& 0 < \alpha \le w(x, t) \text{ for a.e. } (x, t) \in \Omega_T
\end{aligned}
\right\}.
\end{equation}
$W_0$ is a nonempty, closed, bounded, and convex subset of the Banach space $W(0,T)$. 
Consider the mapping
\[
\mathcal{P} : W_0 \rightarrow W_0, \quad w \mapsto I_w,
\]
where $I_w$ is the unique solution to the linearized problem. To ensure that $\mathcal{P}$ maps $W_0$ into itself, we choose $M$ sufficiently large so as to satisfy the energy estimates in Lemma \ref{lemma:linear_combined}. Thanks to Lemma  \ref{lemma:linear_combined}, one can use classical results of compact inclusion in Sobolev spaces to extract subsequences $\{w_{k_n}\}$ of $\{w_k\}$ and $\{I_{k_n}\}$ of $\{I_k\}$ such that for some $I \in \mathcal{W}_0$, the following convergences hold as $k \rightarrow \infty$:
\begin{equation} \label{eq:convergences_list}
\left\{
\begin{aligned}
& w_k \rightarrow w && \text{in } L^2(0, T; L^2(\Omega)) \text{ and a.e. on } \Omega_T, \\
& G_\sigma \ast w_k \rightarrow G_\sigma \ast w && \text{in } L^2(0, T; L^2(\Omega)) \text{ and a.e. on } \Omega_T, \\
& |G_\sigma \ast w_k|^\nu \rightarrow |G_\sigma \ast w|^\nu && \text{in } L^2(0, T; L^2(\Omega)) \text{ and a.e. on } \Omega_T, \\
& \frac{2|G_\sigma \ast w_k|^\nu}{(M_{w_k}^\sigma)^\nu + |G_\sigma \ast w_k|^\nu} \rightarrow \frac{2|G_\sigma \ast w|^\nu}{(M_w^\sigma)^\nu + |G_\sigma \ast w|^\nu} && \text{in } L^2(0, T; L^2(\Omega)) \text{ and a.e. on } \Omega_T, \\
& \Delta(G_\sigma \ast w_k) \rightarrow \Delta(G_\sigma \ast w) && \text{in } L^2(0, T; L^2(\Omega)) \text{ and a.e. on } \Omega_T, \\
& \frac{1}{1 + \left( \frac{|\Delta G_\sigma \ast w_k|}{K} \right)^2} \rightarrow \frac{1}{1 + \left( \frac{|\Delta G_\sigma \ast w|}{K} \right)^2} && \text{in } L^2(0, T; L^2(\Omega)) \text{ and a.e. on } \Omega_T, \\
& \mathcal{S}(w_k) \rightarrow \mathcal{S}(w) && \text{strongly in } L^2(\Omega_T) \text{ and a.e. on } \Omega_T, \\
& I_k \rightharpoonup I && \text{weakly-* in } L^\infty(0, T; H^2(\Omega)), \\
& I_k \rightarrow I && \text{in } L^2(0, T; L^2(\Omega)), \\
& \partial_t I_k \rightharpoonup \partial_t I && \text{weakly-* in } L^\infty(0, T; L^2(\Omega)), \\
& \partial_{tt} I_k \rightharpoonup \partial_{tt} I && \text{weakly in } L^2(0, T; (H^2(\Omega))').
\end{aligned}
\right.
\end{equation}

The convergences established in \eqref{eq:convergences_list}  allow us to pass to the limit in the linearized weak formulation. Let $\{I_k\}$ be the sequence of weak solutions to \eqref{eq:linearized_pde} corresponding to $\{w_k\}$, i.e., $I_k = \mathcal{P}(w_k)$. For any test function $\phi \in H^2$, we have:
\begin{equation} \label{eq:weak_k}
\int_0^T \int_{\Omega}  \partial_{tt}I_{k} \phi \, dx dt + \gamma \int_0^T \int_{\Omega} \partial_{t}I_{k} \phi \, dx dt + \int_0^T \int_{\Omega} \bar{g}(w_k) \Delta I_k \Delta \phi \, dx dt + \int_0^T \int_{\Omega} \mathcal{S}(w_k) \phi \, dx dt = 0.
\end{equation}
From the weak convergences $ \partial_{tt}I_{k} \rightharpoonup \partial_{tt}I$ and $ \partial_{t}I_{k} \rightharpoonup \partial_{t}I$ in their respective spaces, we have:
\begin{equation} \label{eq:lim_linear_terms}
    \int_0^T \int_{\Omega} \left(\partial_{tt}I_{k} + \gamma \partial_{t}I_{k}\right) \phi \, dx dt \longrightarrow \int_0^T \int_{\Omega} (I_{tt} + \gamma I_{t}) \phi \, dx dt, \quad \text{as } k \to \infty.
\end{equation}
Considering the nonlinear diffusion term, we add and subtract $\bar{g}(w) \Delta I_k \Delta \phi$ and apply the triangle inequality to obtain:
\begin{align} \label{eq:weak_derivative}
    &\left| \int_0^T \int_{\Omega} \left( \bar{g}(w_k) \Delta I_k - \bar{g}(w) \Delta I \right) \Delta \phi \, dx dt \right| \nonumber \\
    &\quad \le \int_0^T \int_{\Omega} |\bar{g}(w_k) - \bar{g}(w)| |\Delta I_k| |\Delta \phi| \, dx dt + \left| \int_0^T \int_{\Omega} \bar{g}(w) (\Delta I_k - \Delta I) \Delta \phi \, dx dt \right|.
\end{align}
Where,
 \begin{equation} \label{eq:frozen_coeffs_1}
\bar{g}(w_k):= C(\bar{w_k},_\sigma, |\Delta \bar{w_k},_\sigma|) \quad.
\end{equation}
For the first term on the right-hand side of \eqref{eq:weak_derivative}, we apply the Cauchy-Schwarz inequality, the Lipschitz continuity of $\bar{g}$ (Lemma 2.7), and the uniform $L^2$-bound of $\Delta I_k$:
\begin{equation*}
    \int_0^T \int_{\Omega} |\bar{g}(w_k) - \bar{g}(w)| |\Delta I_k| |\Delta \phi| \, dx dt \le C \|\Delta \phi\|_{L^\infty(\Omega_T)} \|\Delta I_k\|_{L^2(\Omega_T)} \|w_k - w\|_{L^2(\Omega_T)},
\end{equation*}
which converges to $0$ as $k \to \infty$ because $w_k \to w$ strongly in $L^2(\Omega_T)$. 

For the second term, $\Delta I_k \rightharpoonup \Delta I$ weakly in $L^2(\Omega_T)$, the second term also naturally converges to $0$ as $k \to \infty$. Thus, the entire difference in \eqref{eq:weak_derivative} vanishes in the limit.
For the source term, and by the Lebesgue Dominated Convergence Theorem, we have:
\begin{equation} \label{eq:lim_source}
    \int_0^T \int_{\Omega} \mathcal{S}(w_k) \phi \, dx dt \longrightarrow \int_0^T \int_{\Omega} \mathcal{S}(w) \phi \, dx dt, \quad \text{as } k \to \infty.
\end{equation}
By putting the value of \eqref{eq:weak_derivative},\eqref{eq:lim_source} in  \eqref{eq:weak_k} the equation \eqref{eq:weak_k} become:
\begin{equation} \label{eq:weak_limit}
\int_0^T \int_{\Omega} I_{tt} \phi \, dx dt + \gamma \int_0^T \int_{\Omega} I_t \phi \, dx dt + \int_0^T \int_{\Omega} \bar{g}(w) \Delta I \Delta \phi \, dx dt + \int_0^T \int_{\Omega} \mathcal{S}(w) \phi \, dx dt = 0,
\end{equation}
which implies $I = \mathcal{P}(w)$. By the uniqueness of the solution to the linearized problem, the entire sequence $I_k = \mathcal{P}(w_k)$ converges weakly in $W_0$ to $I = \mathcal{P}(w)$. Hence, the mapping $\mathcal{P}$ is weakly continuous from $W_0$ into $W_0$.

This weak continuity, combined with the compactness of the operator, allows us to apply the Schauder fixed-point theorem. We conclude that there exists a fixed point $w \in W_0$ such that $w = \mathcal{P}(w) = I_w$, which solves the original nonlinear problem \eqref{eq:main_pde} to \eqref{eq:bc_ic}.
\vspace{0.2cm}

\subsection*{Uniqueness of the weak solution.} Under Assumption 2.2, we prove the uniqueness of the weak solution for the fourth-order telegraph-diffusion problem \eqref{eq:main_pde}--\eqref{eq:bc_ic}. Let $I_1$ and $I_2$ be two weak solutions of \eqref{eq:main_pde}--\eqref{eq:bc_ic} in the space $W(0, T)$. Then, we have

\begin{equation}
I_{tt} + \gamma I_t + \Delta(C_1 \Delta I) = -\Delta([C_1 - C_2]\Delta I_2) - (\mathcal{S}(I_1) - \mathcal{S}(I_2)) \label{eq:uniqueness_diff}
\end{equation}
subject to the homogeneous initial and boundary conditions:
\begin{equation}
I(x, 0) = 0, \quad I_t(x, 0) = 0, \quad \partial_n I = \partial_n (\Delta I) = 0. \label{eq:bc_ic_diff}
\end{equation}
Here, $C_i \equiv C(I_{\sigma,i}, |\Delta I_{\sigma,i}|)$ and $\mathcal{S}(I_i) \equiv \lambda \left(\frac{I_i - f}{I_i}\right)^2$ for $i=1,2$.

To show $I \equiv 0$, fix $0 < s < T$ and define the backward-integrated variable for the difference:
\begin{equation}
v(\cdot, t) = \begin{cases} \int_t^s I(\cdot, \tau) d\tau, & 0 < t \leq s \\ 0, & s \le t < T \end{cases} \label{eq:v_def}
\end{equation}
Note that for $t \in (0,T)$, we have $\partial_t v(x,t) = -I(x,t)$, and $v(\cdot, s) = 0$. 

Multiplying the difference equation \eqref{eq:uniqueness_diff} by $v$, integrating over $\Omega \times (0, s)$, and using integration by parts along with the Cauchy-Schwarz inequality, we analyze the terms. For the diffusion term, since $\Delta I = -\partial_t \Delta v$, we utilize the following 
:
\begin{equation}
C_1 (-\partial_t \Delta v) \Delta v = -\frac{1}{2} \partial_t \left( C_1 |\Delta v|^2 \right) + \frac{1}{2} \partial_t C_1 |\Delta v|^2.
\end{equation}
Integrating this expression over the space-time domain $\Omega \times (0, s)$ yields:
\begin{align*}
\int_0^s \int_\Omega \Delta(C_1 \Delta I) v \, dx dt &= \int_0^s \int_\Omega C_1 \Delta I \Delta v \, dx dt = -\int_0^s \int_\Omega C_1 (\partial_t \Delta v) \Delta v \, dx dt \\
&= -\frac{1}{2} \int_\Omega \int_0^s \partial_t (C_1 |\Delta v|^2) \, dt dx + \frac{1}{2} \int_0^s \int_\Omega \partial_t C_1 |\Delta v|^2 \, dx dt \\
&= -\frac{1}{2} \int_\Omega \left[ C_1 |\Delta v|^2 \right]_0^s dx + \frac{1}{2} \int_0^s \int_\Omega \partial_t C_1 |\Delta v|^2 \, dx dt.
\end{align*}
We have $v(x, s) = 0$, which implies $\Delta v(x, s) = 0$. Consequently, the boundary term at $t=s$ vanishes, leaving:
\begin{equation*}
-\frac{1}{2} \int_\Omega \left[ C_1(x, s) |\Delta v(x, s)|^2 - C_1(x, 0) |\Delta v(x, 0)|^2 \right] dx = \frac{1}{2} \int_\Omega C_1(x, 0) |\Delta v(x, 0)|^2 dx.
\end{equation*}
Since $v(x,s) = 0$, the boundary term yields $\frac{1}{2} \int_\Omega C_1(x,0) |\Delta v(x,0)|^2 dx$. Combining this with the time derivative terms, we obtain the energy inequality:
\begin{align}
\frac{1}{2} \| I(s) \|^2_{L^2} &+ \gamma \int_0^s \| I(t) \|^2_{L^2} dt + \frac{1}{2} \int_\Omega C_1(x, 0) |\Delta v(x, 0)|^2 dx \nonumber \\
&\leq \frac{1}{2} \left| \int_0^s \int_\Omega |\Delta v|^2 \partial_t C_1 \, dx dt \right| + \int_0^s \| C_1 - C_2 \|_{L^\infty} \| \Delta I_2 \|_{L^2} \| \Delta v \|_{L^2} dt \nonumber \\
&\quad + \left| \int_0^s \int_\Omega (\mathcal{S}(I_1) - \mathcal{S}(I_2)) v \, dx dt \right|. \label{eq:v_energy}
\end{align}
Since both $I_1$ and $I_2$ are weak solutions in the admissible set $W_0$, they satisfy the pointwise lower bound $I_i(x,t) \ge \alpha > 0$. Therefore, the source term is locally Lipschitz continuous with an adaptive constant $L_{\mathcal{S}}$. Using Cauchy-Schwarz and Young's inequalities, we obtain:
\begin{align}
\left| \int_0^s \int_\Omega (\mathcal{S}(I_1) - \mathcal{S}(I_2)) v \, dx dt \right| &\le L_{\mathcal{S}} \int_0^s \|I(t)\|_{L^2} \|v(t)\|_{L^2} dt \nonumber \\
&\le \frac{L_{\mathcal{S}}}{2} \int_0^s \left( \|I(t)\|_{L^2}^2 + \|v(t)\|_{L^2}^2 \right) dt. \label{eq:source_lipschitz}
\end{align}

Using lemma \ref{lemma:coeff_prop} and the coercivity $C_1(x,0) \ge \kappa > 0$, we substitute the Lipschitz bounds \eqref{eq:source_lipschitz} into the energy inequality \eqref{eq:v_energy}. Also noting that $\| \partial_t C_1 \|_{L^\infty} \leq \tilde{C}$, we apply Young's inequality to obtain:
\begin{equation}
\frac{1}{2} \| I(s) \|^2_{L^2} + \int_0^s \| I(t) \|^2_{L^2} dt + C \| \Delta v(0) \|^2_{L^2} \leq C \int_0^s \left( \| \Delta v(t) \|^2_{L^2} + \| I(t) \|^2_{L^2} + \| v(t) \|^2_{L^2} \right) dt. \label{eq:energy_v_bound}
\end{equation}

Notice that $\| v(t) \|^2_{L^2} = \| \int_t^s I(\tau) d\tau \|^2_{L^2} \leq T \int_0^s \| I(\tau) \|^2_{L^2} d\tau$. This allows us to absorb $\|v(t)\|_{L^2}^2$ into the integral of $I(t)$. 

Now, we introduce the forward integral variable $w(x, t) = \int_0^t I(x, \tau) d\tau$. Observe that $v(x, 0) = \int_0^s I(\tau) d\tau = w(x, s)$. For our fourth-order problem, the $L^2$-norm of the Laplacian $\|\Delta w\|_{L^2}$ controls the $H^2$-norm. Replacing $v(0)$ with $w(s)$, equation \eqref{eq:energy_v_bound} reduces to:
\begin{equation}
\frac{1}{2} \|I(s)\|_{L^2}^2 + \int_0^s \|I(t)\|_{L^2}^2 dt + C \|w(s)\|_{H^2}^2 \le \tilde{C}s \|w(s)\|_{H^2}^2 + C \int_0^s \left( \|w(t)\|_{H^2}^2 + \|I(t)\|_{L^2}^2 \right) dt. \label{eq:energy_w}
\end{equation}

To absorb the problematic term $\tilde{C}s \|w(s)\|_{H^2}^2$ into the left-hand side, we restrict the time interval. Choose $T_1$ sufficiently small such that $C - \tilde{C}T_1 > 0$. Then, for any $0 < s \le T_1$, we can subtract the term from both sides, absorbing it into the coercivity constant. Hence, \eqref{eq:energy_w} reduces to:
\begin{equation}
\|I(s)\|_{L^2}^2 + \|w(s)\|_{H^2}^2 \le C \int_0^s \left( \|w(t)\|_{H^2}^2 + \|I(t)\|_{L^2}^2 \right) dt. \label{eq:gronwall_ready}
\end{equation}

Defining the total energy $E(s) = \|I(s)\|_{L^2}^2 + \|w(s)\|_{H^2}^2$, the inequality simplifies to:
\begin{equation}
E(s) \le C \int_0^s E(t) dt \quad \text{for } 0 < s \le T_1. \label{eq:final_gronwall}
\end{equation}
An application of Grönwall's lemma implies $I \equiv 0$ almost everywhere on $[0, T_1]$. 
Finally, we utilize similar logic on the successive intervals $(T_1, 2T_1], (2T_1, 3T_1], \dots$ step by step. Since the length of the interval $T_1$ depends only on the universal constants $C$ and $\tilde{C}$, we eventually deduce that $I_1 = I_2$ on the entire interval $(0, T)$. This completes the proof of the theorem.

\section{Numerical Implementation}\label{sec:discrtization}
We consider the proposed fourth-order PDE as given in equation (28):

\begin{equation}
\frac{\partial^2 I}{\partial t^2} + \gamma \frac{\partial I}{\partial t} = -\Delta \left( C(I_\xi, |\Delta I_\xi|) \Delta I\right) - \lambda \left( \frac{I - f}{I} \right)^2, \quad \forall (x, t) \in \Omega \times (0, T),
\end{equation}

The 2D domain \( \Omega \) is discretized into a grid with spatial step sizes \( \delta x \), \( \delta y \), and a temporal step size \( \delta t \). Let \( I^n_{i,j} \) represent the numerical approximation of \( I \) at grid point \((x_i, y_j, t_n)\). We consider the following discretization: 

\[
\frac{I^{n+1}_{i,j} - 2I^n_{i,j} + I^{n-1}_{i,j}}{(\delta t)^2} + \gamma \frac{I^{n+1}_{i,j} - I^n_{i,j}}{\delta t} 
= -\Delta \left( C(I^n_{\xi, i, j}, |\Delta I^n_{\xi, i, j}|) \Delta I^n_{ i, j} \right) 
- \lambda \left( \frac{I^n_{i,j} - f_{i,j}}{I^n_{i,j}} \right)^2.
\]
Here temporal discretization employs a second-order central difference for the second derivative, 
\[
I_{tt}(t_n) \approx \frac{I^{n+1}_{i,j} - 2I^n_{i,j} + I^{n-1}_{i,j}}{(\delta t)^2},
\]
and a first-order forward difference for the first derivative, 
\[
I_t(t_n) \approx \frac{I^{n+1}_{i,j} - I^n_{i,j}}{\delta t}.
\]
For spatial derivatives, second-order central differences are used for both the gradient,
\[
\nabla_x I^n_{i,j} \approx \frac{I^n_{i+1,j} - I^n_{i-1,j}}{2\delta x}, \quad 
\nabla_y I^n_{i,j} \approx \frac{I^n_{i,j+1} - I^n_{i,j-1}}{2\delta y},
\]
and the Laplacian,
\[
\Delta I^n_{i,j} \approx \frac{I^n_{i+1,j} - 2I^n_{i,j} + I^n_{i-1,j}}{(\delta x)^2} 
+ \frac{I^n_{i,j+1} - 2I^n_{i,j} + I^n_{i,j-1}}{(\delta y)^2}.
\]
   The laplacian magnitude(\(|\Delta I|\)) is computed as:
   \[
   |\Delta I^n_{i,j}| = \sqrt{\left( \Delta_x I^n_{i,j} \right)^2 + \left( \Delta_y I^n_{i,j} \right)^2}.
   \]

Rearranging terms to solve for \( I^{n+1}_{i,j} \), we obtain:

\[
I^{n+1}_{i,j} = \frac{1}{(\delta t)^2 + \gamma \delta t} 
\Bigg[ 2I^n_{i,j} - I^{n-1}_{i,j} 
- (\delta t)^2 \Delta \left( C(I^n_{\xi, i, j}, |\Delta I^n_{\xi, i, j}|) \Delta I^n_{ i, j} \right)
- (\delta t)^2 \lambda  \left( \frac{I^n_{i,j} - f_{i,j}}{I^n_{i,j}} \right)^2 \Bigg].
\]

The discretized fidelity term \( \lambda \frac{(I^n_{i,j})^2 - (f_{i,j})^2}{(I^n_{i,j})^2} \) gives a balance between preservation of the image's original features and noise reduction.
For color images, the restoration process is applied separately to each channel (Red, Green, and Blue). Let \( I^n_{i,j,s} \) denote the intensity at pixel \( (i,j) \) in channel \( s \) at time step \( n \), where \( s \in \{R, G, B\} \). The discretized equation for each channel follows:
\begin{equation}
\frac{I^{n+1}_{i,j,s} - 2I^n_{i,j,s} + I^{n-1}_{i,j,s}}{(\delta t)^2} + \gamma \frac{I^{n+1}_{i,j,s} - I^n_{i,j,s}}{\delta t} 
= -\Delta \left( C(I^n_{\xi, i, j, s}, |\Delta I^n_{\xi, i, j, s}|) \Delta I^n_{ i, j, s} \right) 
- \lambda \left( \frac{I^n_{i,j,s} - f_{i,j,s}}{I^n_{i,j,s}} \right)^2.
\end{equation}
Each channel is processed independently using the same numerical scheme as in the grayscale case. Spatial derivatives (\(\nabla I, \Delta I\)) and the fidelity term are computed separately for each channel. The final restored color image is obtained by merging the individually denoised channels:

\begin{equation}
I_{\text{denoised}} = \text{cat}(I_{\text{denoised}, R}, I_{\text{denoised}, G}, I_{\text{denoised}, B}).
\end{equation}

\section{ Numerical Experiments and Performance Evaluation}

In this section the proposed model in compared with existing second-order Shan\cite{shan2019smooth} and TDM\cite{Majee} using grayscale images (Boat, Texture), two images from the BSD68 dataset, and color images (Baboon, Peppers), all corrupted with multiplicative speckle noise at varying intensity levels ($L = {1, 3, 5, 10}$). The noise was generated using MATLAB's \texttt{gamrnd} function as $\eta = \text{gamrnd}(L, 1/L, A, B)$, where $A \times B$ represents the image dimensions.
Experiments were conducted in MATLAB on an Intel Core i7 (3.60 GHz) with 8 GB RAM. The proposed model was evaluated across five images, ensuring consistency by applying the same finite-difference discretization across all methods.
For numerical stability, the explicit scheme adhered to the CFL condition \cite{zauderer2011partial,li2009numerical,zhang2014tensor}, given by $\tau \leq \frac{h}{\max g(x,t)}$, where $h$ is the spatial step size. The time steps of $\tau = 0.2$ and $\xi = 2$ were used. PSNR was used as the convergence criterion for natural images to ensure optimal restoration quality, whereas for SAR images, relative error was used.

\begin{figure}[H]
    \centering
    \begin{tabular}{cccc}
        \includegraphics[width=0.22\textwidth]{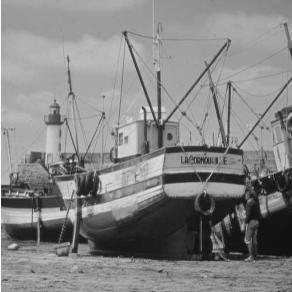} &
        \hspace{-5pt}\includegraphics[width=0.22\textwidth]{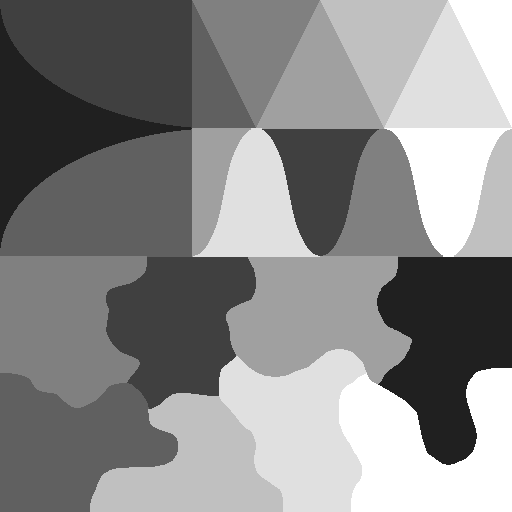} &
        \hspace{-5pt}\includegraphics[width=0.22\textwidth]{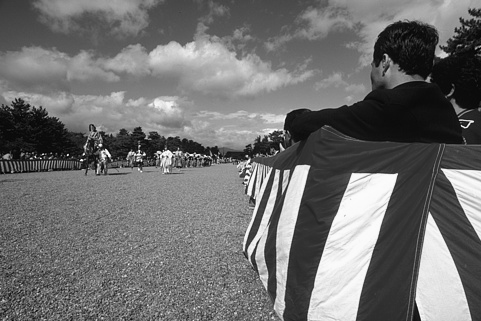} &
        \hspace{-5pt}\includegraphics[width=0.22\textwidth]{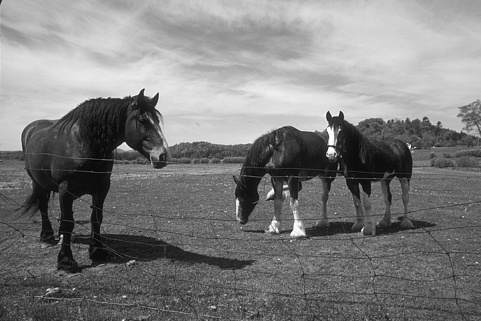} \\
        (a) Boat & (b) Texture & (c) Test 019 & (d) Test 037 
    \end{tabular}
    \caption{Grayscale test images: (a) Boat, (b) Texture, (c) Test 019, and (d) Test 037.}
    \label{fig:grayscale_row}
\end{figure}

\begin{figure}[H]
\centering
\begin{tabular}{cccc} 
\includegraphics[width=0.25\textwidth]{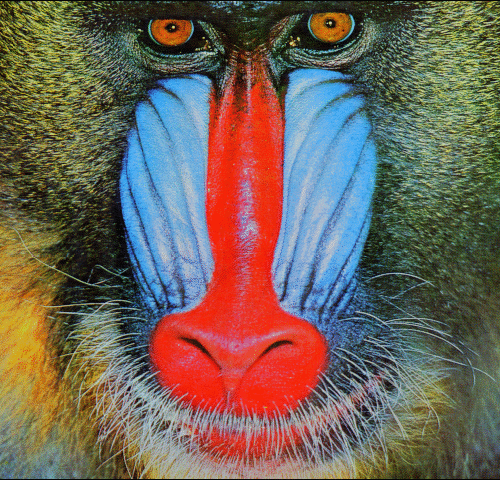} & 
\includegraphics[width=0.24\textwidth]{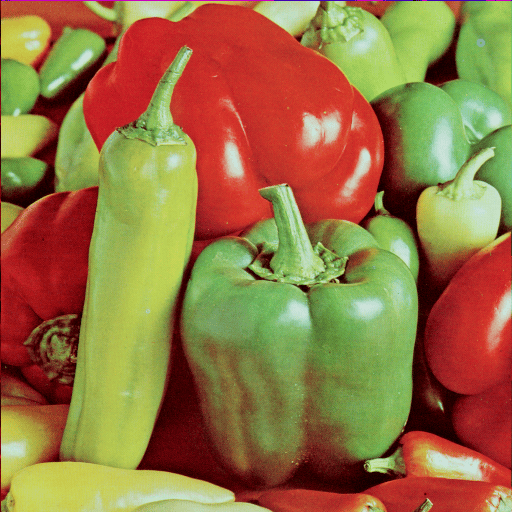}  
\end{tabular}
\caption{color Images: (a) Baboon, (b)Pepper.}
\label{fig:images2}
\end{figure}

For comparison, we used the following test images: a grayscale Boat image of size $292 \times 292$, two images from the BSD68 data set, each of size $321 \times 481$, and a Texture image of size $512 \times 512$ (Figure~\ref{fig:grayscale_row}); two color images, namely Baboon of size $500 \times 480$ and Peppers of size $512 \times 512$ (Figure~\ref{fig:images2}). 
For synthetic aperture radar (SAR) images, the Speckle Index is used to measure noise reduction. The SI is given by:
\begin{equation}
SI = \frac{\text{S.D.}(I)}{\text{M}(I)},
\end{equation}
where $\text{S.D.}(I)$ and $\text{M}(I)$ represent the standard deviation and mean intensities of the pixels, respectively.
The PSNR is calculated as:
\begin{equation}
\text{PSNR} = 10 \cdot \log_{10} \left( \frac{\text{MAX}^2}{\text{MSE}} \right),
\end{equation}
where MSE represents the Mean Squared Error, and MAX is the maximum pixel value.
The Structural Similarity Index (SSIM) and MSSIM are defined as:
\begin{equation}
\text{SSIM}(a, b) = \frac{(2\mu_a \mu_b + d_1)(2\sigma_{ab} + d_2)}{(\mu_a^2 + \mu_b^2 + d_1)(\sigma_a^2 + \sigma_b^2 + d_2)}, \quad \text{MSSIM} = \frac{1}{N} \sum_{i=1}^{N} \text{SSIM}_i.
\end{equation}
where $\mu_a, \mu_b$ are mean values, $\sigma_a^2, \sigma_b^2$ are variances, and $\sigma_{ab}$ is the covariance. The constants $d_1$ and $d_2$ stabilize the calculation, and N is the number of iterations.

For SAR image denoising, we apply specific norm-based criteria to evaluate the convergence of the proposed model. The convergence is generally assessed using the relative error norm and the Speckle Index. 
When the ground truth image is unavailable, the relative error norm serves as the stopping criterion. The numerical iteration is terminated when the relative error between two consecutive iterations falls below a specific threshold $\epsilon$, as described in \cite{Majee}. For the BSD68 Dataset experimental comparisons employed each method’s individually optimized parameter settings to ensure unbiased evaluation. Specifically, Shan’s model\cite{shan2019smooth} was run with $\alpha=0.1$, $\beta=1$; the TDM model\cite{Majee} with $\gamma=5$, $\alpha=1$, $k=2$; and the proposed model with $\gamma=5$, $\alpha=2$, $k=2$, and $\lambda=0.08$.

\begin{figure}[H]
    \centering
    \begin{subfigure}[b]{0.24\textwidth}
        \centering
        \includegraphics[width=\textwidth]{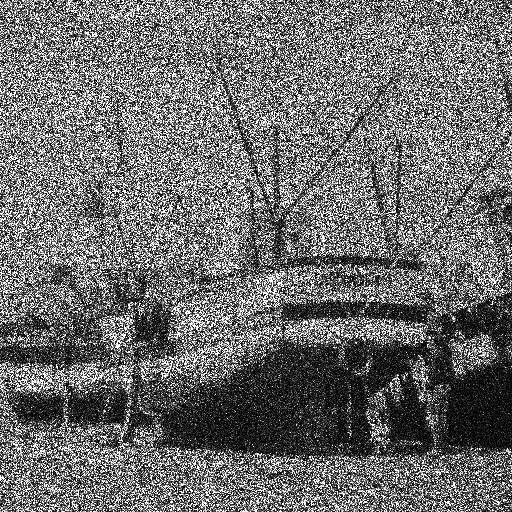}
        \caption{Look=1}
    \end{subfigure}
    \begin{subfigure}[b]{0.24\textwidth}
        \centering
        \includegraphics[width=\textwidth]{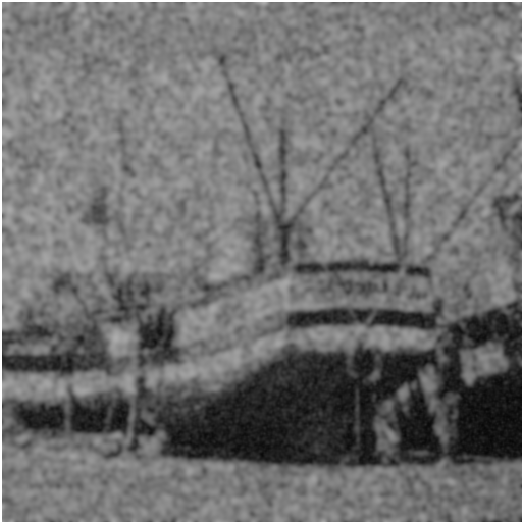}
        \caption{SHAN\cite{shan2019smooth}}
    \end{subfigure}
    \begin{subfigure}[b]{0.24\textwidth}
        \centering
        \includegraphics[width=\textwidth]{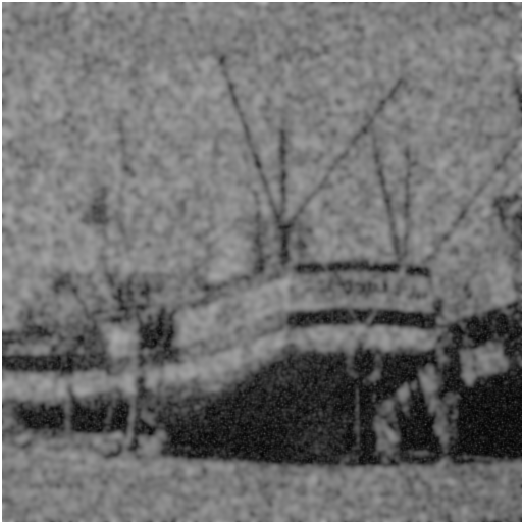}
        \caption{TDM\cite{Majee}}
    \end{subfigure}
    \begin{subfigure}[b]{0.24\textwidth}
        \centering
        \includegraphics[width=\textwidth]{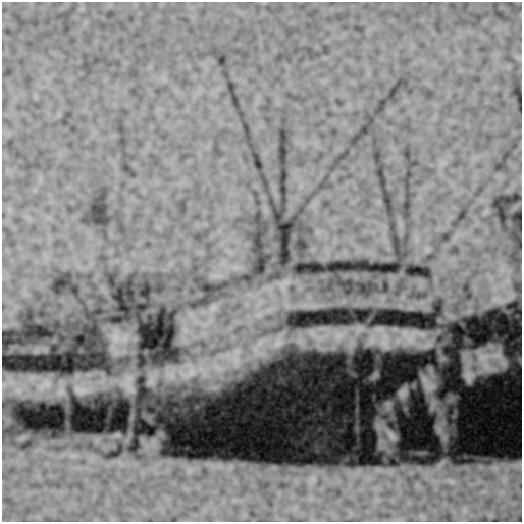}
        \caption{Proposed}
    \end{subfigure}

    \begin{subfigure}[b]{0.24\textwidth}
        \centering
        \includegraphics[width=\textwidth]{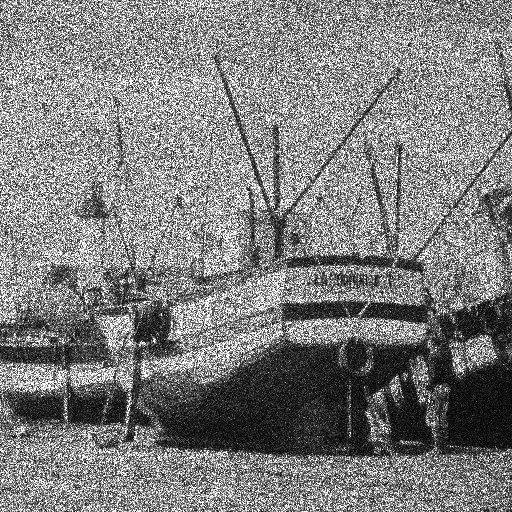}
        \caption{Look=3}
    \end{subfigure}
    \begin{subfigure}[b]{0.24\textwidth}
        \centering
        \includegraphics[width=\textwidth]{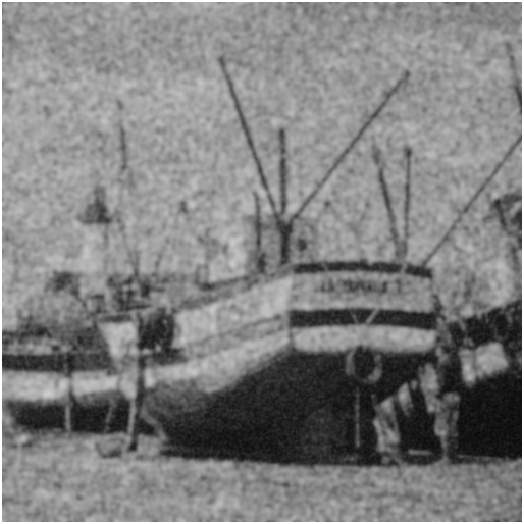}
        \caption{SHAN\cite{shan2019smooth}}
    \end{subfigure}
    \begin{subfigure}[b]{0.24\textwidth}
        \centering
        \includegraphics[width=\textwidth]{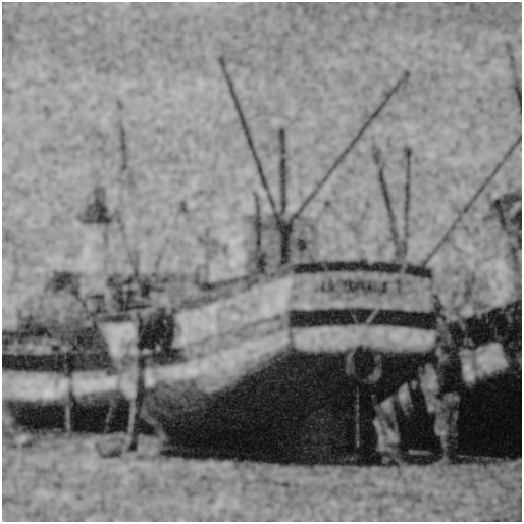}
        \caption{TDM\cite{Majee}}
    \end{subfigure}
    \begin{subfigure}[b]{0.24\textwidth}
        \centering
        \includegraphics[width=\textwidth]{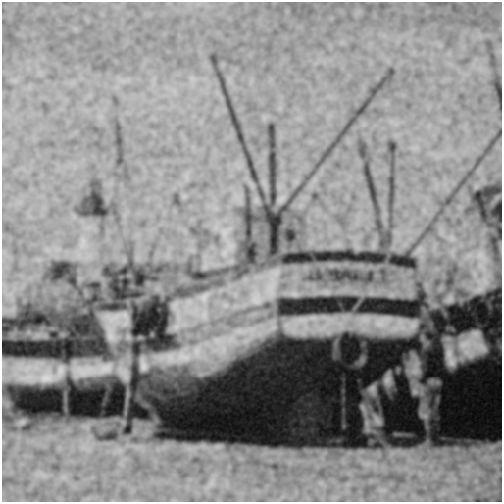}
        \caption{Proposed}
    \end{subfigure}

    \begin{subfigure}[b]{0.24\textwidth}
        \centering
        \includegraphics[width=\textwidth]{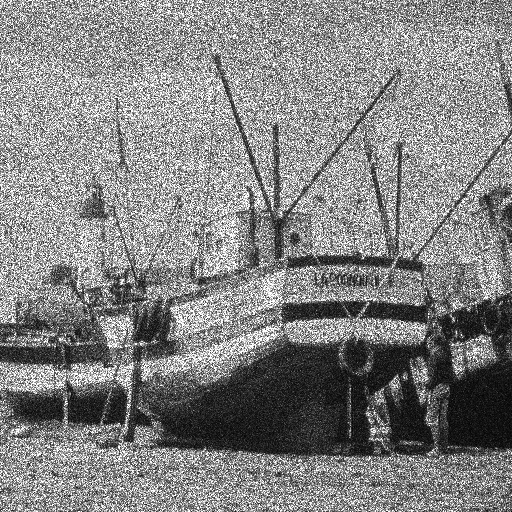}
        \caption{Look=5}
    \end{subfigure}
    \begin{subfigure}[b]{0.24\textwidth}
        \centering
        \includegraphics[width=\textwidth]{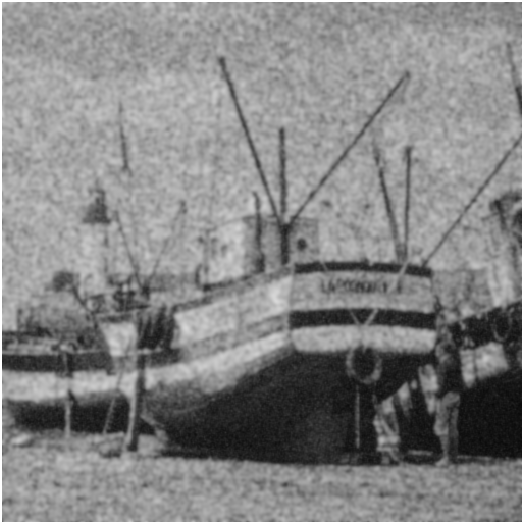}
        \caption{SHAN\cite{shan2019smooth}}
    \end{subfigure}
    \begin{subfigure}[b]{0.24\textwidth}
        \centering
        \includegraphics[width=\textwidth]{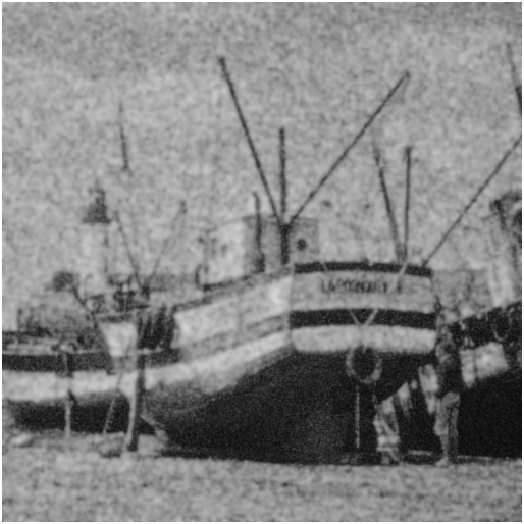}
        \caption{TDM\cite{Majee}}
    \end{subfigure}
    \begin{subfigure}[b]{0.24\textwidth}
        \centering
        \includegraphics[width=\textwidth]{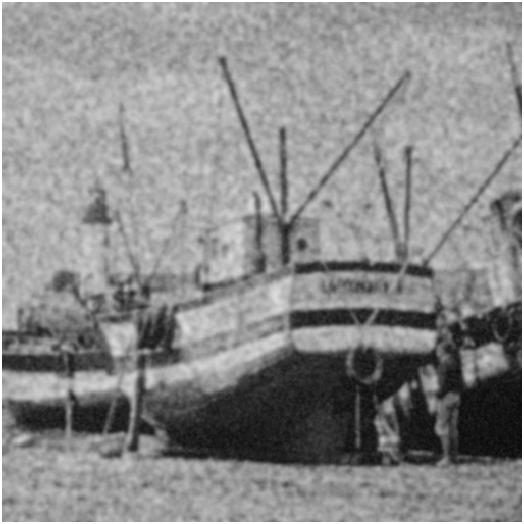}
        \caption{Proposed}
    \end{subfigure}

    \begin{subfigure}[b]{0.24\textwidth}
        \centering
        \includegraphics[width=\textwidth]{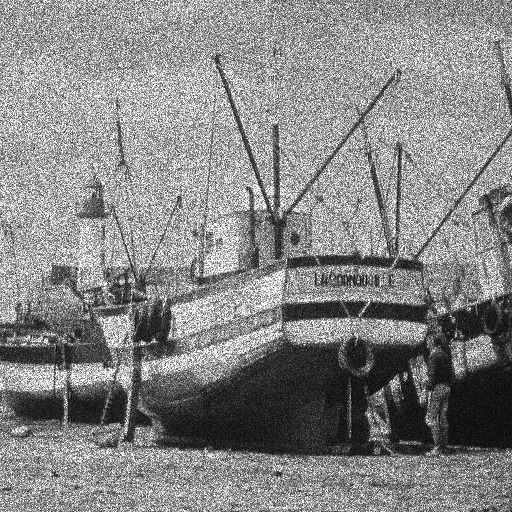}
        \caption{Look=10}
    \end{subfigure}
    \begin{subfigure}[b]{0.24\textwidth}
        \centering
        \includegraphics[width=\textwidth]{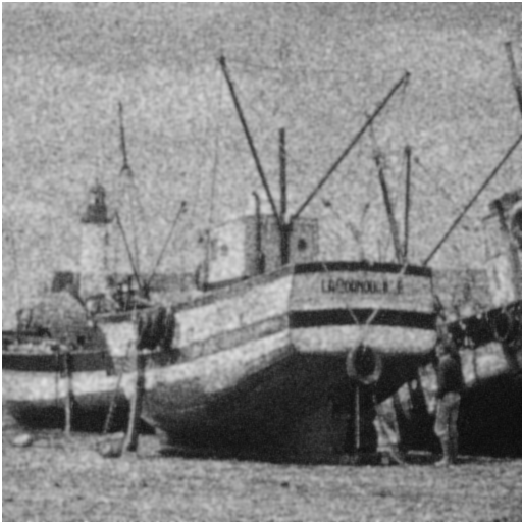}
        \caption{SHAN\cite{shan2019smooth}}
    \end{subfigure}
    \begin{subfigure}[b]{0.24\textwidth}
        \centering
        \includegraphics[width=\textwidth]{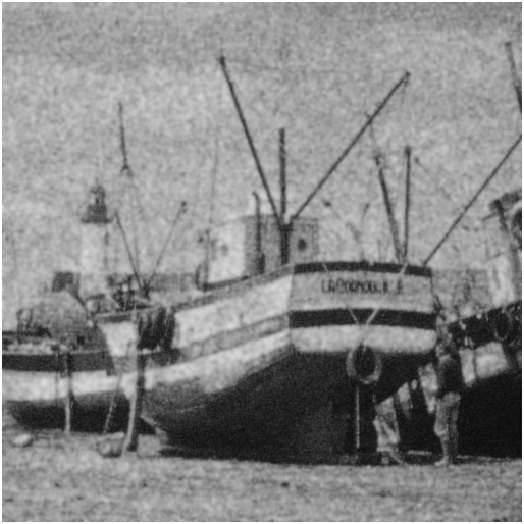}
        \caption{TDM\cite{Majee}}
    \end{subfigure}
    \begin{subfigure}[b]{0.24\textwidth}
        \centering
        \includegraphics[width=\textwidth]{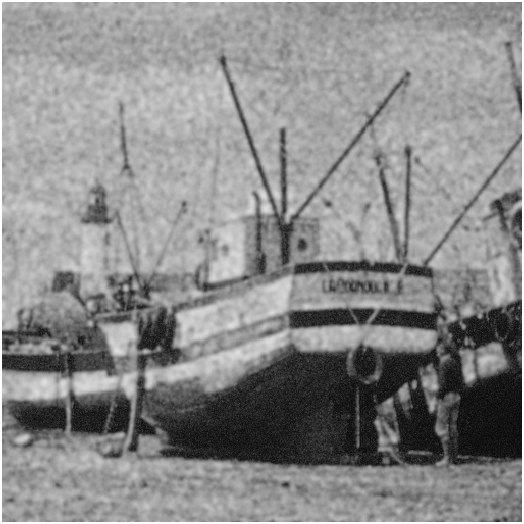}
        \caption{Proposed}
    \end{subfigure}

    \caption{
        The first column contains noisy boat images with noise level Look = 1, 3, 5, 10. 
        Subsequent columns: Restored boat images using different models (SHAN\cite{shan2019smooth}, TDM \cite{Majee}, Proposed).
    }
    \label{fig:gamma_noise_results1}
\end{figure}

\begin{figure}[H]
\centering
    \begin{subfigure}[b]{0.24\textwidth}
        \centering
        \includegraphics[width=\textwidth]{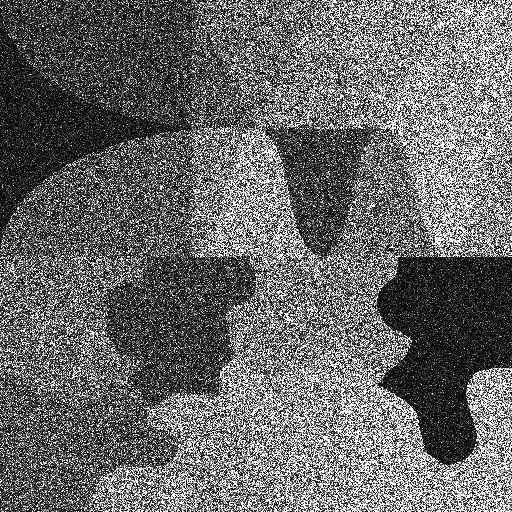}
        \caption{Look=1}
    \end{subfigure}
    \begin{subfigure}[b]{0.24\textwidth}
        \centering
        \includegraphics[width=\textwidth]{shan_texture_Restored_look_1.png}
        \caption{SHAN\cite{shan2019smooth}}
    \end{subfigure}
    \begin{subfigure}[b]{0.24\textwidth}
        \centering
        \includegraphics[width=\textwidth]{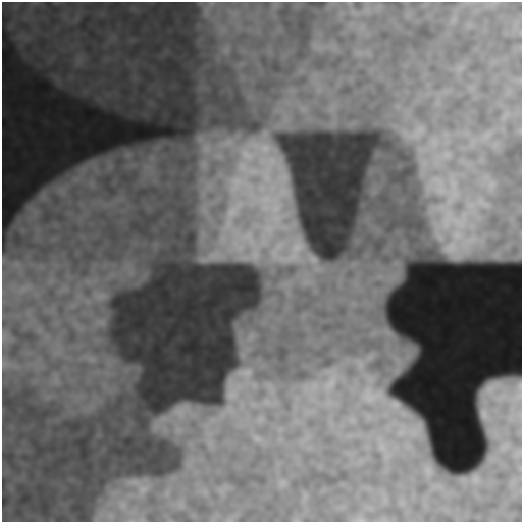}
        \caption{TDM\cite{Majee}}
    \end{subfigure}
    \begin{subfigure}[b]{0.24\textwidth}
        \centering
        \includegraphics[width=\textwidth]{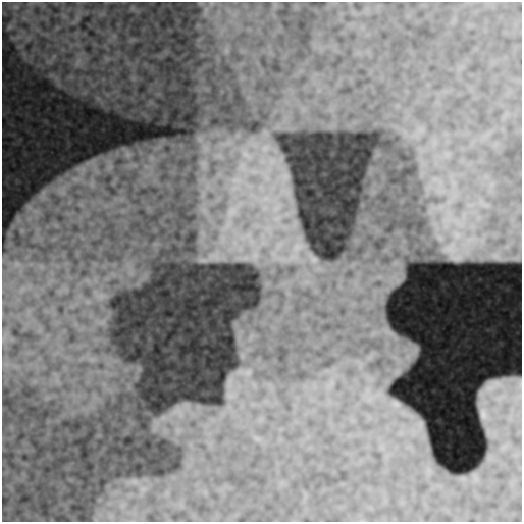}
        \caption{Proposed}
    \end{subfigure}

    \begin{subfigure}[b]{0.24\textwidth}
        \centering
        \includegraphics[width=\textwidth]{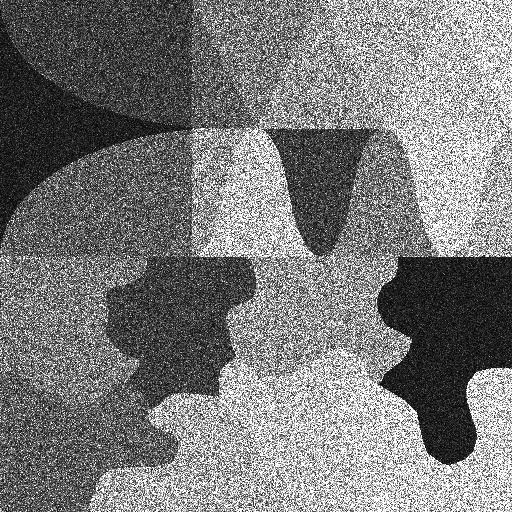}
        \caption{Look=3}
    \end{subfigure}
    \begin{subfigure}[b]{0.24\textwidth}
        \centering
        \includegraphics[width=\textwidth]{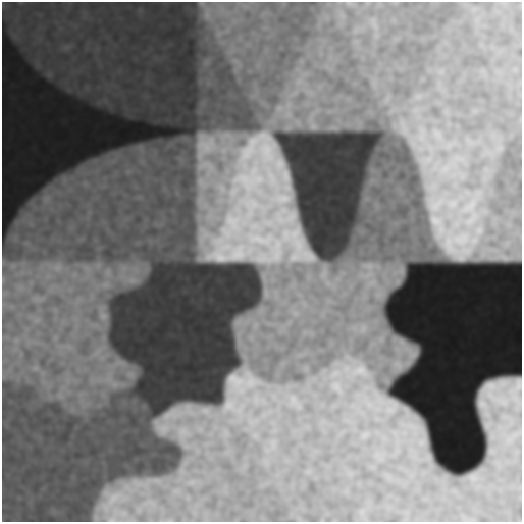}
        \caption{SHAN\cite{shan2019smooth}}
    \end{subfigure}
    \begin{subfigure}[b]{0.24\textwidth}
        \centering
        \includegraphics[width=\textwidth]{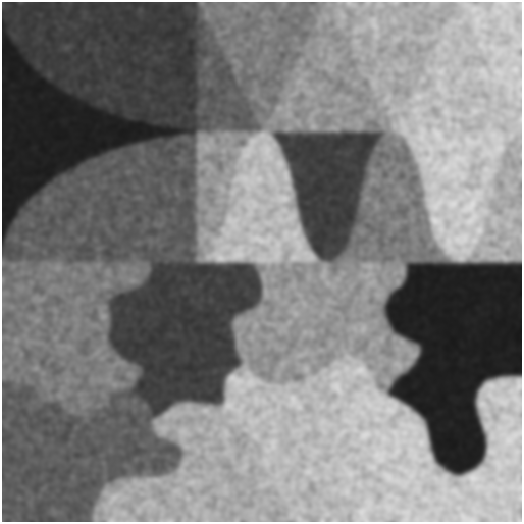}
        \caption{TDM\cite{Majee}}
    \end{subfigure}
    \begin{subfigure}[b]{0.24\textwidth}
        \centering
        \includegraphics[width=\textwidth]{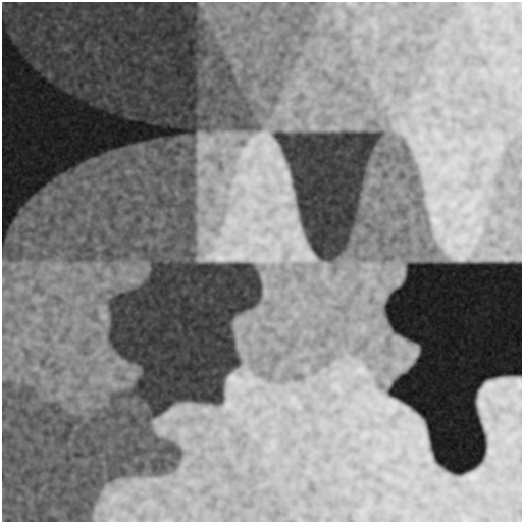}
        \caption{Proposed}
    \end{subfigure}

    \begin{subfigure}[b]{0.24\textwidth}
        \centering
        \includegraphics[width=\textwidth]{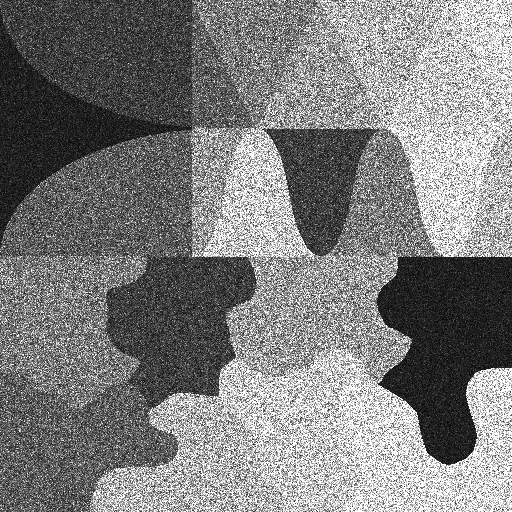}
        \caption{Look=5}
    \end{subfigure}
    \begin{subfigure}[b]{0.24\textwidth}
        \centering
        \includegraphics[width=\textwidth]{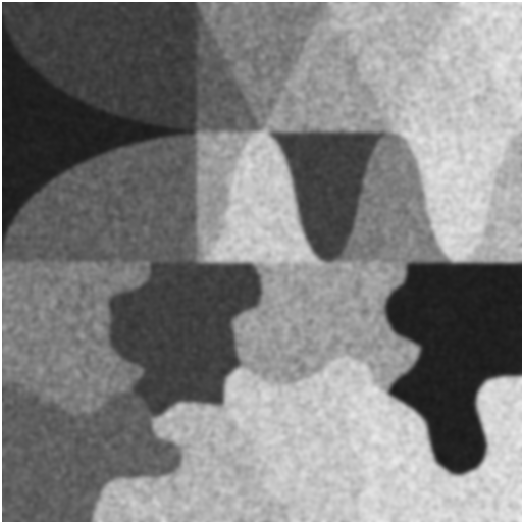}
        \caption{SHAN\cite{shan2019smooth}}
    \end{subfigure}
    \begin{subfigure}[b]{0.24\textwidth}
        \centering
        \includegraphics[width=\textwidth]{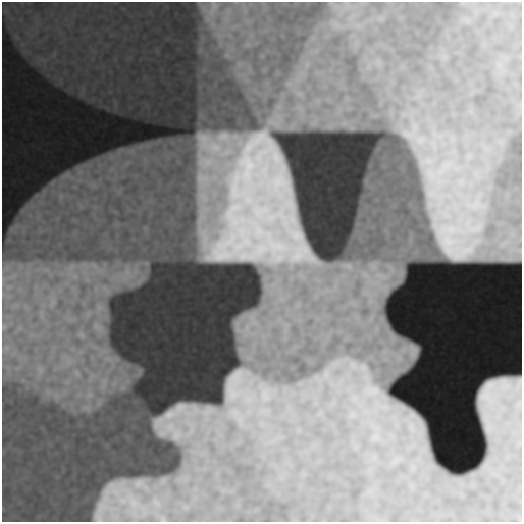}
        \caption{TDM\cite{Majee}}
    \end{subfigure}
    \begin{subfigure}[b]{0.24\textwidth}
        \centering
        \includegraphics[width=\textwidth]{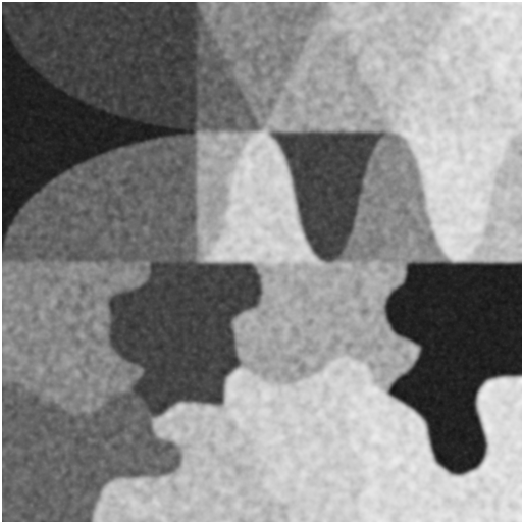}
        \caption{Proposed}
    \end{subfigure}

    \begin{subfigure}[b]{0.24\textwidth}
        \centering
        \includegraphics[width=\textwidth]{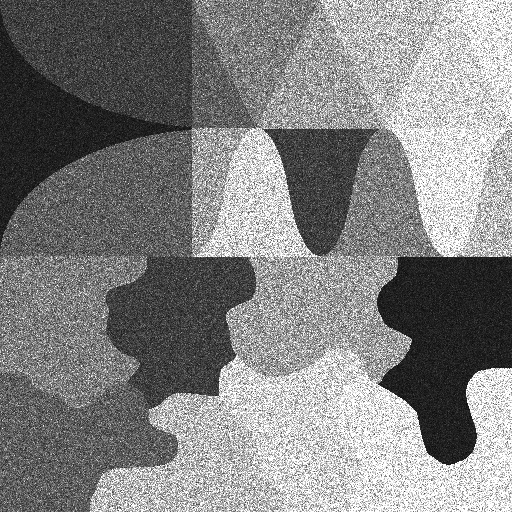}
        \caption{Look=10}
    \end{subfigure}
    \begin{subfigure}[b]{0.24\textwidth}
        \centering
        \includegraphics[width=\textwidth]{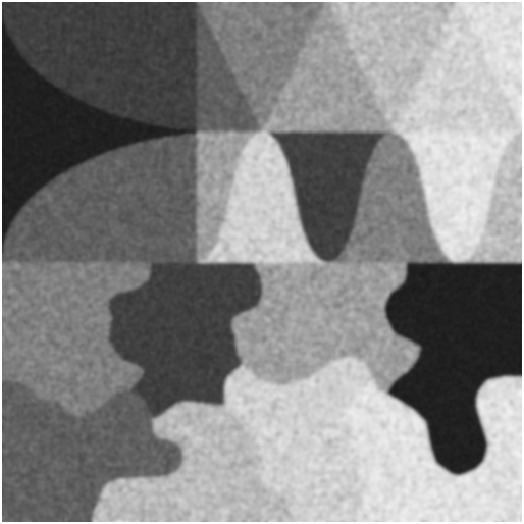}
        \caption{SHAN\cite{shan2019smooth}}
    \end{subfigure}
    \begin{subfigure}[b]{0.24\textwidth}
        \centering
        \includegraphics[width=\textwidth]{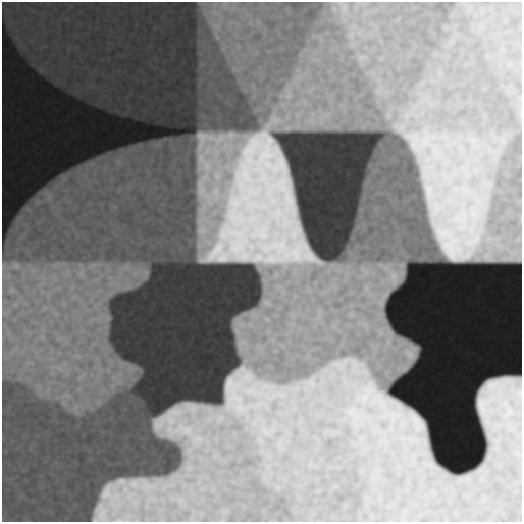}
        \caption{TDM\cite{Majee}}
    \end{subfigure}
    \begin{subfigure}[b]{0.24\textwidth}
        \centering
        \includegraphics[width=\textwidth]{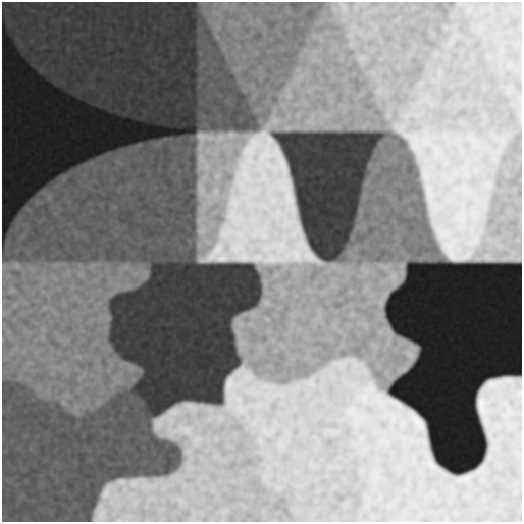}
        \caption{Proposed}
    \end{subfigure}

    \caption{
    The first column contains noisy texture images with noise levels Look = 1, 3, 5, 10. 
    Subsequent columns: Restored texture images using different models (SHAN\cite{shan2019smooth}, TDM\cite{Majee}, and Proposed Model).
    }
    \label{fig:gamma_noise_results2}
\end{figure}

\begin{figure}[H]
\centering
    \begin{subfigure}[b]{0.24\textwidth}
        \centering
        \includegraphics[width=\textwidth]{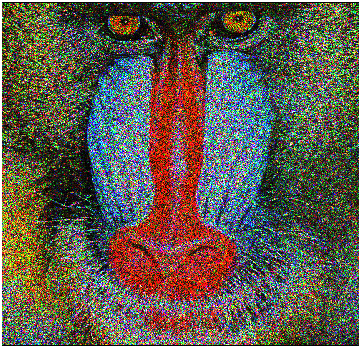}
        \caption{Look=1}
    \end{subfigure}
    \begin{subfigure}[b]{0.24\textwidth}
        \centering
        \includegraphics[width=\textwidth]{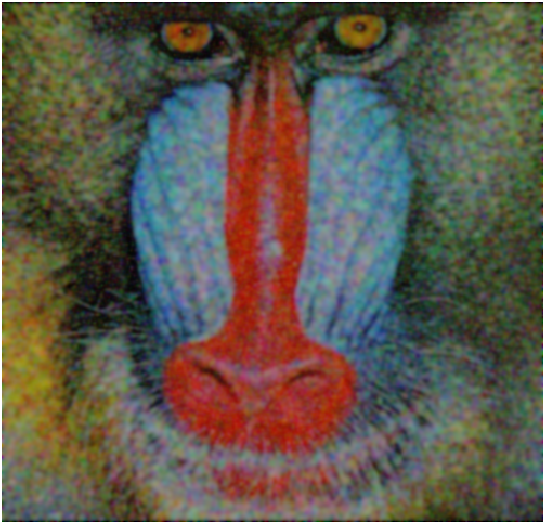}
        \caption{SHAN\cite{shan2019smooth}}
    \end{subfigure}
    \begin{subfigure}[b]{0.24\textwidth}
        \centering
        \includegraphics[width=\textwidth]{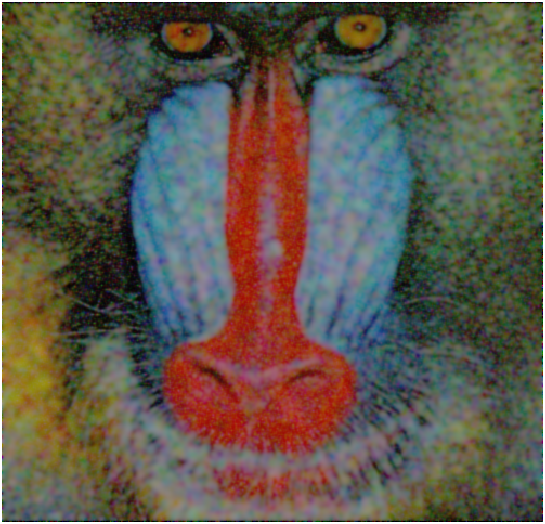}
        \caption{TDM\cite{Majee}}
    \end{subfigure}
    \begin{subfigure}[b]{0.24\textwidth}
        \centering
        \includegraphics[width=\textwidth]{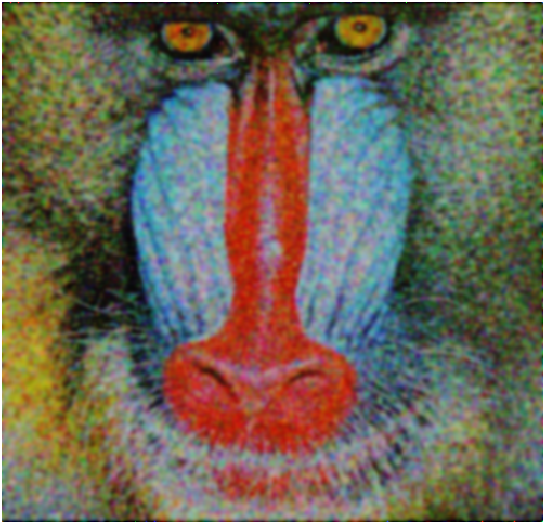}
        \caption{Proposed}
    \end{subfigure}

    \begin{subfigure}[b]{0.24\textwidth}
        \centering
        \includegraphics[width=\textwidth]{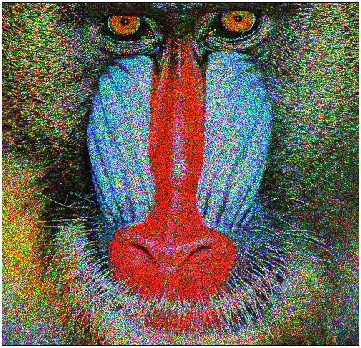}
        \caption{Look=3}
    \end{subfigure}
    \begin{subfigure}[b]{0.24\textwidth}
        \centering
        \includegraphics[width=\textwidth]{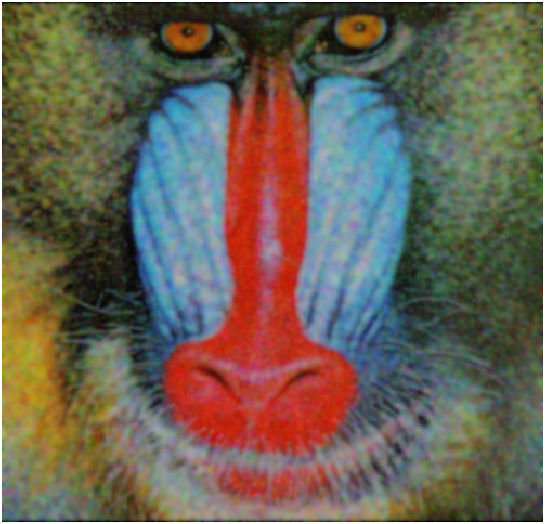}
        \caption{SHAN\cite{shan2019smooth}}
    \end{subfigure}
    \begin{subfigure}[b]{0.24\textwidth}
        \centering
        \includegraphics[width=\textwidth]{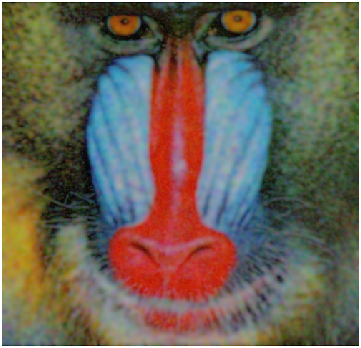}
        \caption{TDM\cite{Majee}}
    \end{subfigure}
    \begin{subfigure}[b]{0.24\textwidth}
        \centering
        \includegraphics[width=\textwidth]{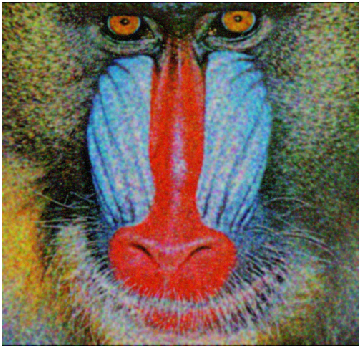}
        \caption{Proposed}
    \end{subfigure}

    \begin{subfigure}[b]{0.24\textwidth}
        \centering
        \includegraphics[width=\textwidth]{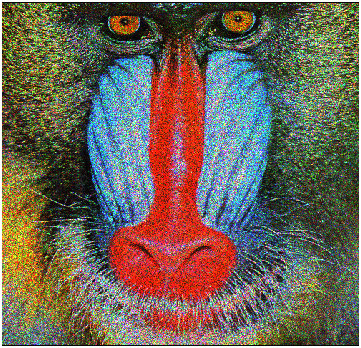}
        \caption{Look=5}
    \end{subfigure}
    \begin{subfigure}[b]{0.24\textwidth}
        \centering
        \includegraphics[width=\textwidth]{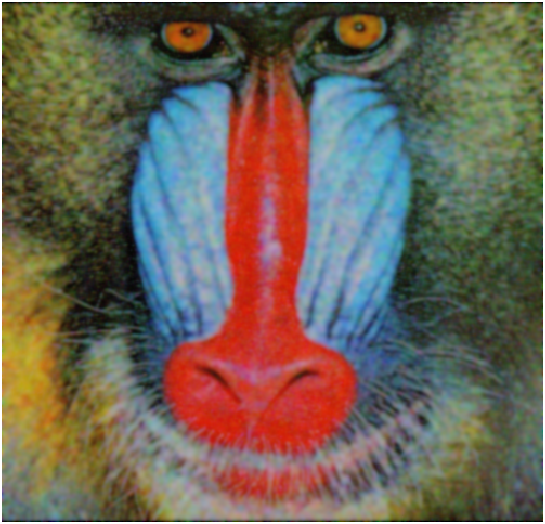}
        \caption{SHAN\cite{shan2019smooth}}
    \end{subfigure}
    \begin{subfigure}[b]{0.24\textwidth}
        \centering
        \includegraphics[width=\textwidth]{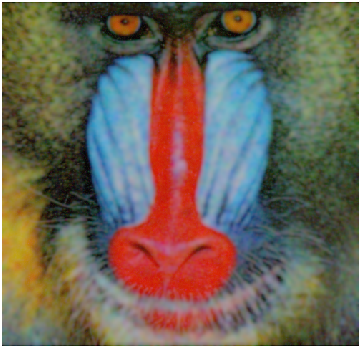}
        \caption{TDM\cite{Majee}}
    \end{subfigure}
    \begin{subfigure}[b]{0.24\textwidth}
        \centering
        \includegraphics[width=\textwidth]{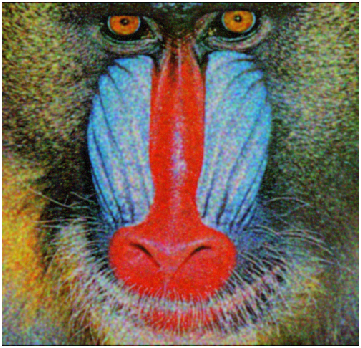}
        \caption{Proposed}
    \end{subfigure}

    \begin{subfigure}[b]{0.24\textwidth}
        \centering
        \includegraphics[width=\textwidth]{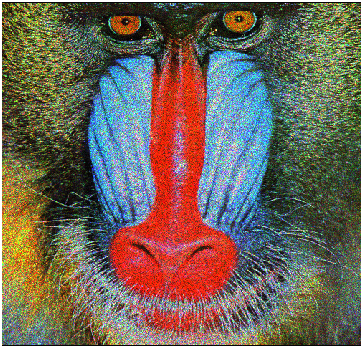}
        \caption{Look=10}
    \end{subfigure}
    \begin{subfigure}[b]{0.24\textwidth}
        \centering
        \includegraphics[width=\textwidth]{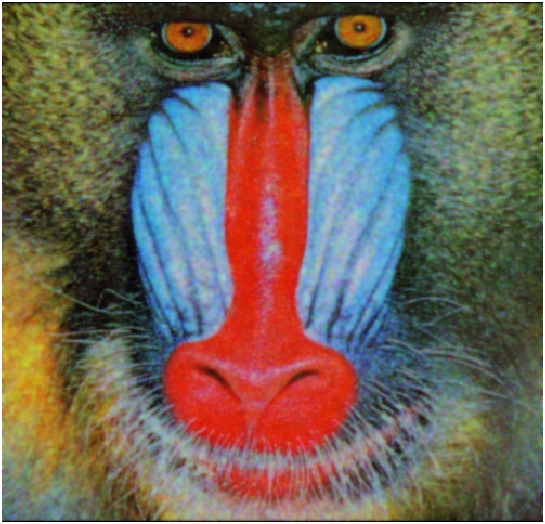}
        \caption{SHAN\cite{shan2019smooth}}
    \end{subfigure}
    \begin{subfigure}[b]{0.24\textwidth}
        \centering
        \includegraphics[width=\textwidth]{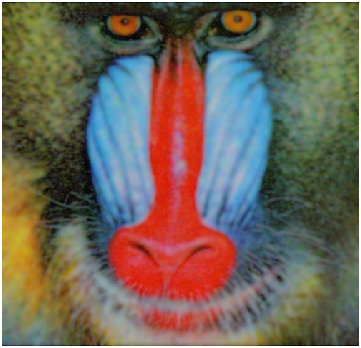}
        \caption{TDM\cite{Majee}}
    \end{subfigure}
    \begin{subfigure}[b]{0.24\textwidth}
        \centering
        \includegraphics[width=\textwidth]{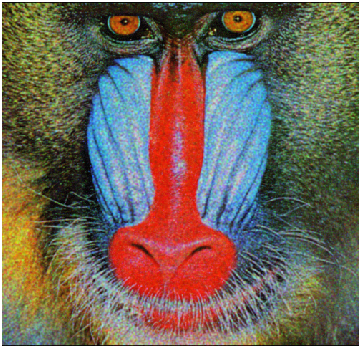}
        \caption{Proposed}
    \end{subfigure}

    \caption{
    The first column contains noisy baboon images with noise levels Look = 1, 3, 5, 10. 
    Subsequent columns: Restored texture images using different models (SHAN\cite{shan2019smooth}, TDM\cite{Majee}, and Proposed Model).
    }
    \label{fig:gamma_noise_results3}
\end{figure}
\begin{figure}[H]
\centering
    \begin{subfigure}[b]{0.24\textwidth}
        \centering
        \includegraphics[width=\textwidth]{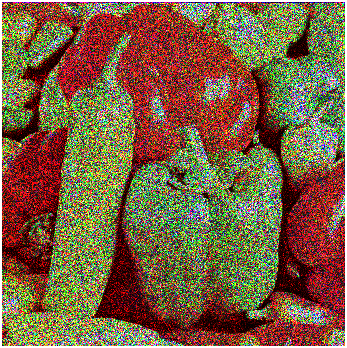}
        \caption{Look=1}
    \end{subfigure}
    \begin{subfigure}[b]{0.24\textwidth}
        \centering
        \includegraphics[width=\textwidth]{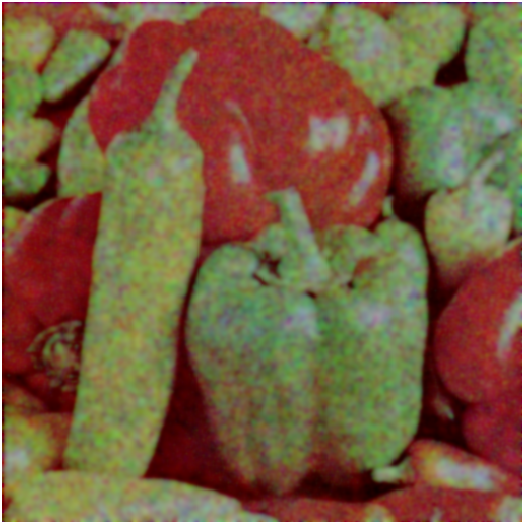}
        \caption{SHAN\cite{shan2019smooth}}
    \end{subfigure}
    \begin{subfigure}[b]{0.24\textwidth}
        \centering
        \includegraphics[width=\textwidth]{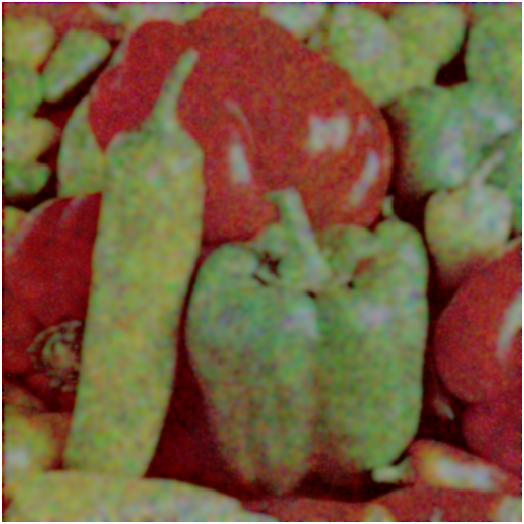}
        \caption{TDM\cite{Majee}}
    \end{subfigure}
    \begin{subfigure}[b]{0.24\textwidth}
        \centering
        \includegraphics[width=\textwidth]{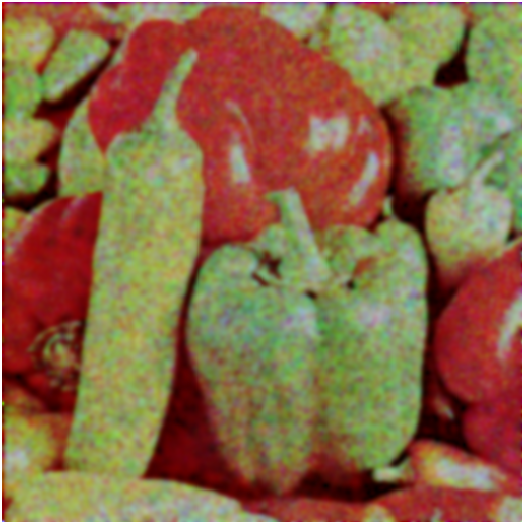}
        \caption{Proposed}
    \end{subfigure}

    \begin{subfigure}[b]{0.24\textwidth}
        \centering
        \includegraphics[width=\textwidth]{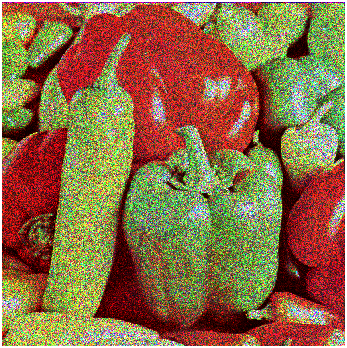}
        \caption{Look=3}
    \end{subfigure}
    \begin{subfigure}[b]{0.24\textwidth}
        \centering
        \includegraphics[width=\textwidth]{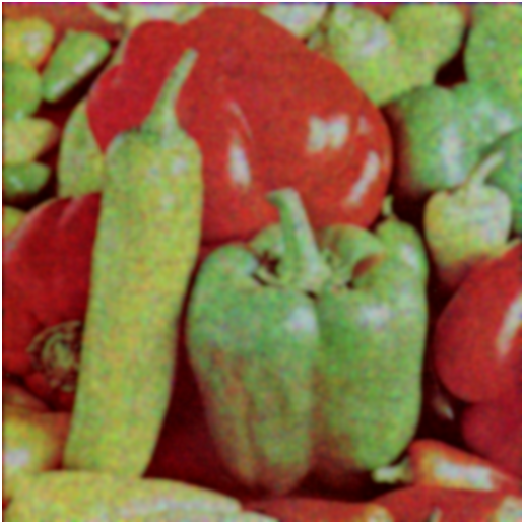}
        \caption{SHAN\cite{shan2019smooth}}
    \end{subfigure}
    \begin{subfigure}[b]{0.24\textwidth}
        \centering
        \includegraphics[width=\textwidth]{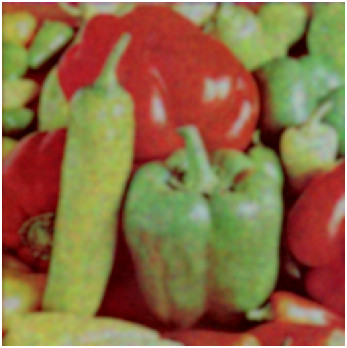}
        \caption{TDM\cite{Majee}}
    \end{subfigure}
    \begin{subfigure}[b]{0.24\textwidth}
        \centering
        \includegraphics[width=\textwidth]{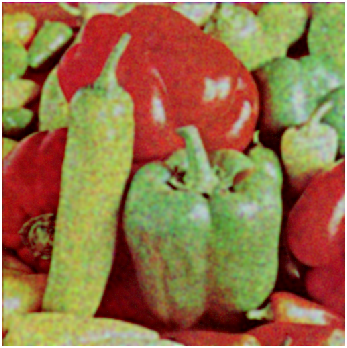}
        \caption{Proposed}
    \end{subfigure}

    \begin{subfigure}[b]{0.24\textwidth}
        \centering
        \includegraphics[width=\textwidth]{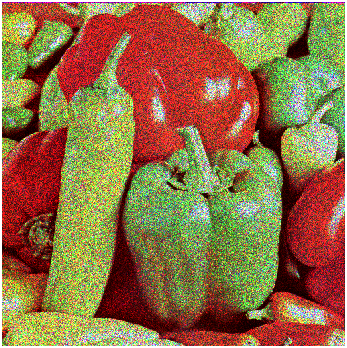}
        \caption{Look=5}
    \end{subfigure}
    \begin{subfigure}[b]{0.24\textwidth}
        \centering
        \includegraphics[width=\textwidth]{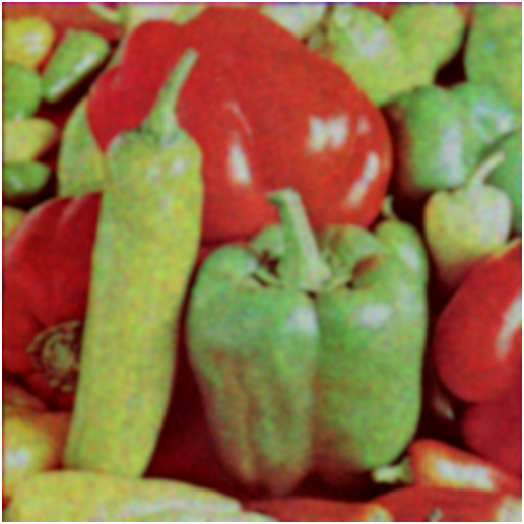}
        \caption{SHAN\cite{shan2019smooth}}
    \end{subfigure}
    \begin{subfigure}[b]{0.24\textwidth}
        \centering
        \includegraphics[width=\textwidth]{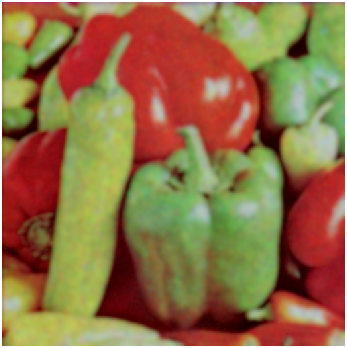}
        \caption{TDM\cite{Majee}}
    \end{subfigure}
    \begin{subfigure}[b]{0.24\textwidth}
        \centering
        \includegraphics[width=\textwidth]{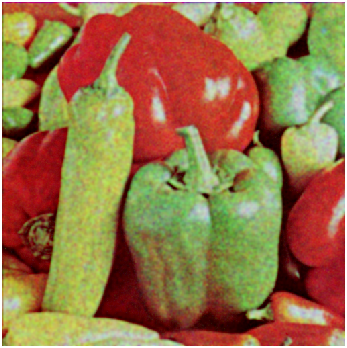}
        \caption{Proposed}
    \end{subfigure}

    \begin{subfigure}[b]{0.24\textwidth}
        \centering
        \includegraphics[width=\textwidth]{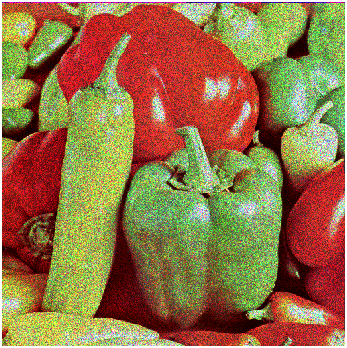}
        \caption{Look=10}
    \end{subfigure}
    \begin{subfigure}[b]{0.24\textwidth}
        \centering
        \includegraphics[width=\textwidth]{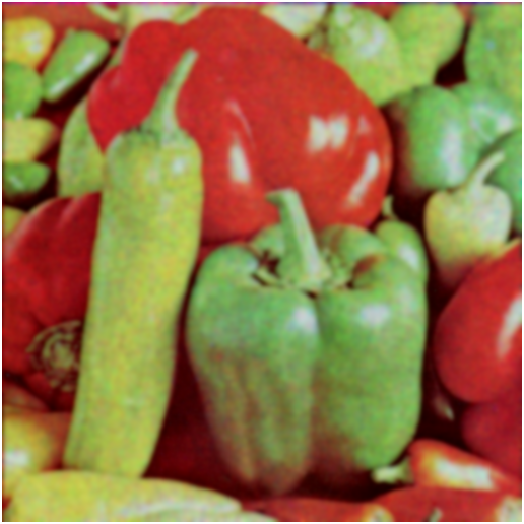}
        \caption{SHAN\cite{shan2019smooth}}
    \end{subfigure}
    \begin{subfigure}[b]{0.24\textwidth}
        \centering
        \includegraphics[width=\textwidth]{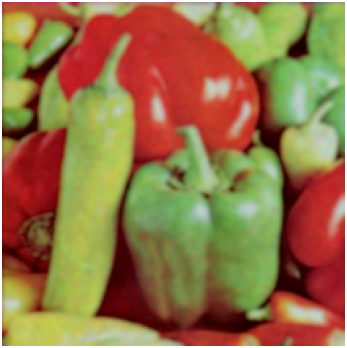}
        \caption{TDM\cite{Majee}}
    \end{subfigure}
    \begin{subfigure}[b]{0.24\textwidth}
        \centering
        \includegraphics[width=\textwidth]{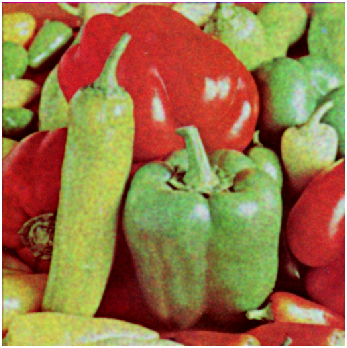}
        \caption{Proposed}
    \end{subfigure}

    \caption{
    The first column contains noisy pepper images with noise levels Look = 1, 3, 5, 10. 
    Subsequent columns: Restored pepper images using different models (SHAN\cite{shan2019smooth}, TDM\cite{Majee}, and Proposed Model).
    }
    \label{fig:gamma_noise_results4}
\end{figure}
\begin{figure}[H]
\centering
    \begin{subfigure}[b]{0.24\textwidth}
        \centering
        \includegraphics[width=\textwidth]{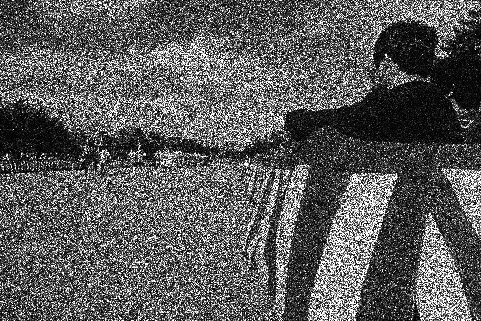}
        \caption{Look=1}
    \end{subfigure}
    \begin{subfigure}[b]{0.24\textwidth}
        \centering
        \includegraphics[width=\textwidth]{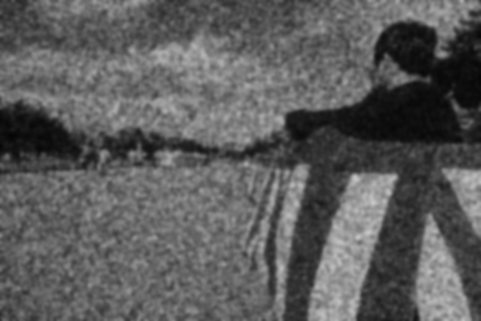}
        \caption{SHAN\cite{shan2019smooth}}
    \end{subfigure}
    \begin{subfigure}[b]{0.24\textwidth}
        \centering
        \includegraphics[width=\textwidth]{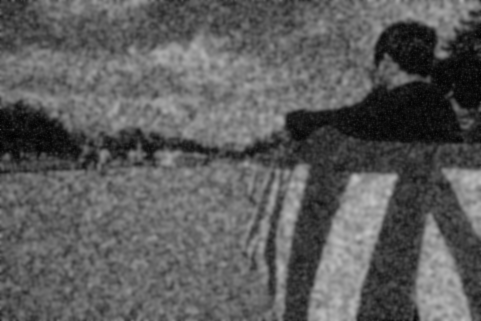}
        \caption{TDM\cite{Majee}}
    \end{subfigure}
    \begin{subfigure}[b]{0.24\textwidth}
        \centering
        \includegraphics[width=\textwidth]{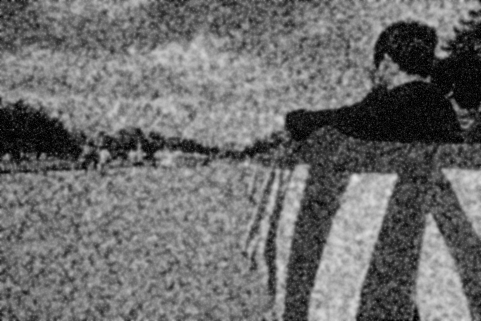}
        \caption{Proposed}
    \end{subfigure}

    \begin{subfigure}[b]{0.24\textwidth}
        \centering
        \includegraphics[width=\textwidth]{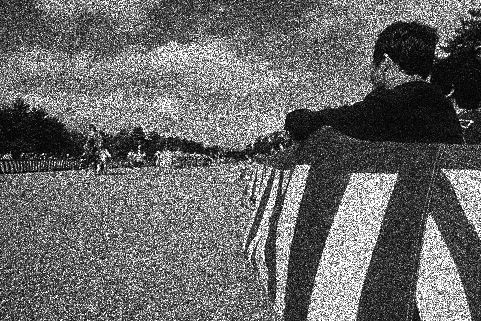}
        \caption{Look=3}
    \end{subfigure}
    \begin{subfigure}[b]{0.24\textwidth}
        \centering
        \includegraphics[width=\textwidth]{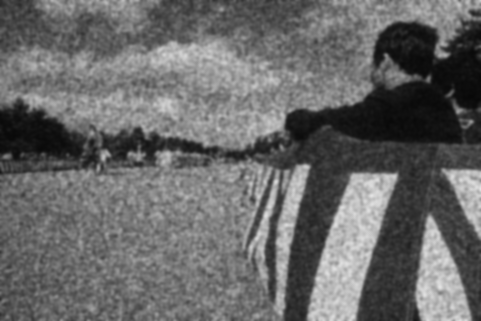}
        \caption{SHAN\cite{shan2019smooth}}
    \end{subfigure}
    \begin{subfigure}[b]{0.24\textwidth}
        \centering
        \includegraphics[width=\textwidth]{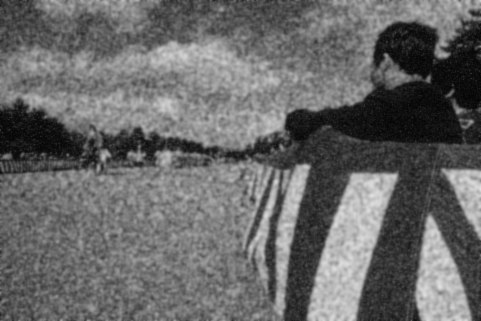}
        \caption{TDM\cite{Majee}}
    \end{subfigure}
    \begin{subfigure}[b]{0.24\textwidth}
        \centering
        \includegraphics[width=\textwidth]{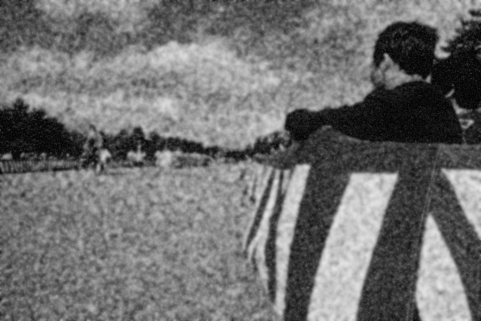}
        \caption{Proposed}
    \end{subfigure}

    \begin{subfigure}[b]{0.24\textwidth}
        \centering
        \includegraphics[width=\textwidth]{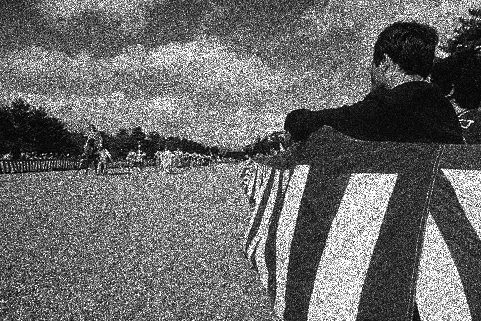}
        \caption{Look=5}
    \end{subfigure}
    \begin{subfigure}[b]{0.24\textwidth}
        \centering
        \includegraphics[width=\textwidth]{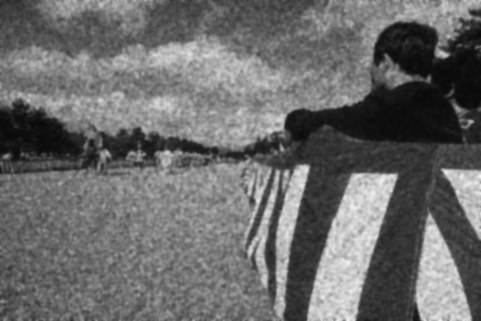}
        \caption{SHAN\cite{shan2019smooth}}
    \end{subfigure}
    \begin{subfigure}[b]{0.24\textwidth}
        \centering
        \includegraphics[width=\textwidth]{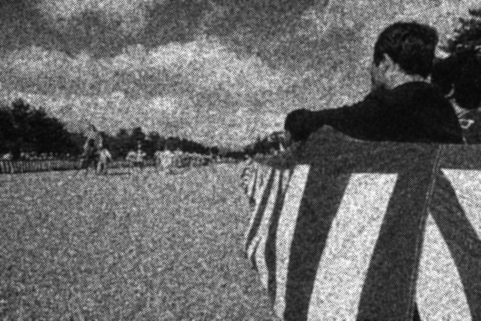}
        \caption{TDM\cite{Majee}}
    \end{subfigure}
    \begin{subfigure}[b]{0.24\textwidth}
        \centering
        \includegraphics[width=\textwidth]{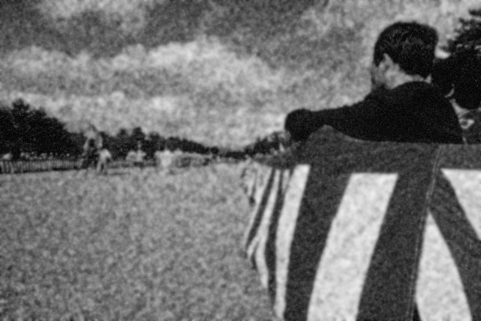}
        \caption{Proposed}
    \end{subfigure}

    \begin{subfigure}[b]{0.24\textwidth}
        \centering
        \includegraphics[width=\textwidth]{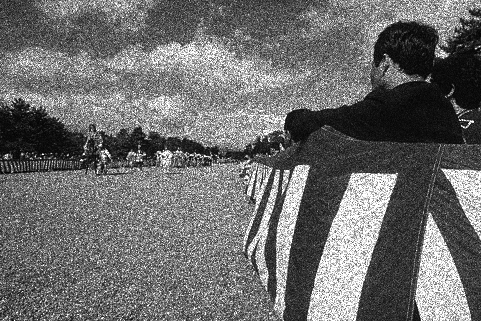}
        \caption{Look=10}
    \end{subfigure}
    \begin{subfigure}[b]{0.24\textwidth}
        \centering
        \includegraphics[width=\textwidth]{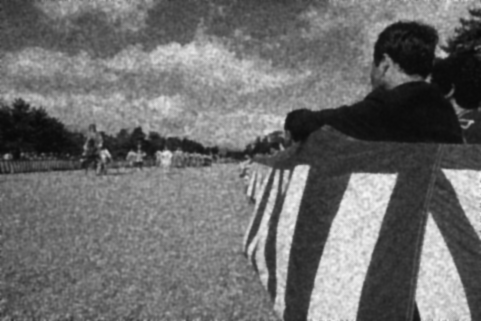}
        \caption{SHAN\cite{shan2019smooth}}
    \end{subfigure}
    \begin{subfigure}[b]{0.24\textwidth}
        \centering
        \includegraphics[width=\textwidth]{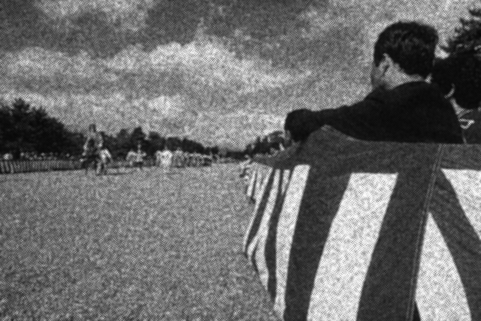}
        \caption{TDM\cite{Majee}}
    \end{subfigure}
    \begin{subfigure}[b]{0.24\textwidth}
        \centering
        \includegraphics[width=\textwidth]{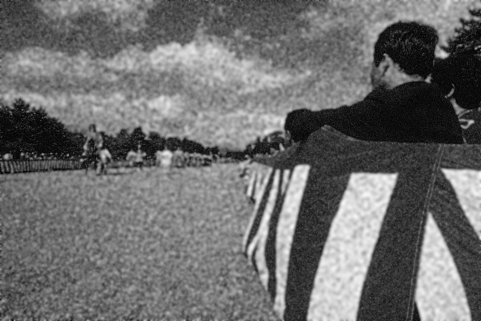}
        \caption{Proposed}
    \end{subfigure}

   \caption{
    The first column contains noisy pepper images with noise levels Look = 1, 3, 5, 10. 
    Subsequent columns: Restored test019 image using different models (SHAN\cite{shan2019smooth}, TDM\cite{Majee}, and Proposed Model).}
    \label{fig:noise_results1}
\end{figure}
\begin{figure}[H]
\centering
    \begin{subfigure}[b]{0.24\textwidth}
        \centering
        \includegraphics[width=\textwidth]{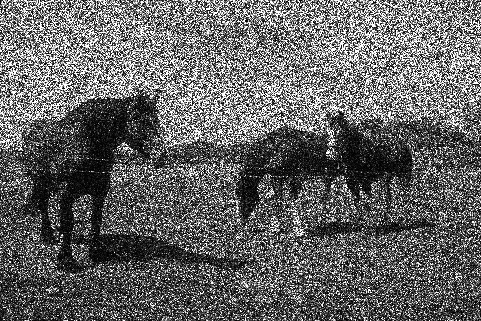}
        \caption{Look=1}
    \end{subfigure}
    \begin{subfigure}[b]{0.24\textwidth}
        \centering
        \includegraphics[width=\textwidth]{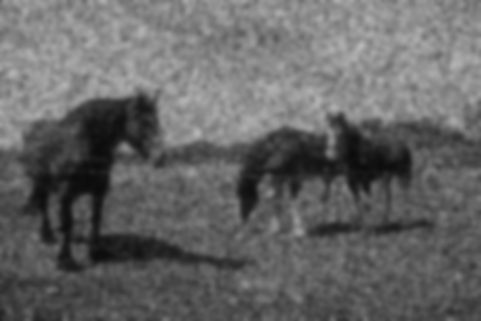}
        \caption{SHAN\cite{shan2019smooth}}
    \end{subfigure}
    \begin{subfigure}[b]{0.24\textwidth}
        \centering
        \includegraphics[width=\textwidth]{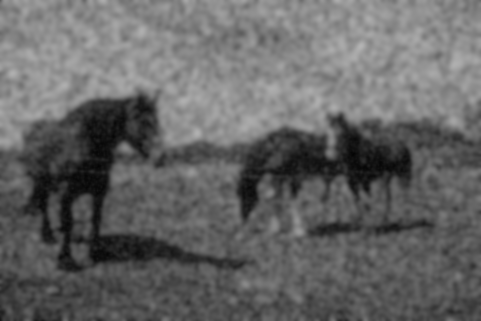}
        \caption{TDM\cite{Majee}}
    \end{subfigure}
    \begin{subfigure}[b]{0.24\textwidth}
        \centering
        \includegraphics[width=\textwidth]{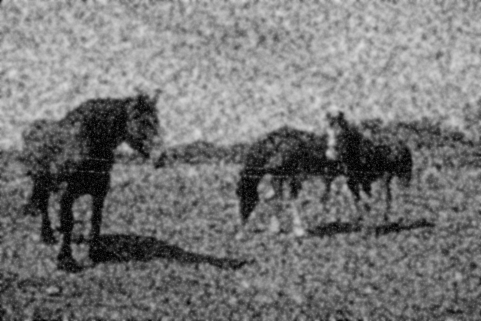}
        \caption{Proposed}
    \end{subfigure}

    \begin{subfigure}[b]{0.24\textwidth}
        \centering
        \includegraphics[width=\textwidth]{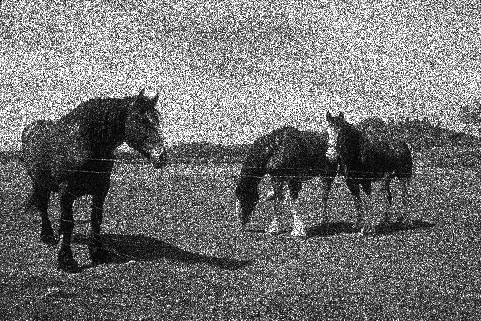}
        \caption{Look=3}
    \end{subfigure}
    \begin{subfigure}[b]{0.24\textwidth}
        \centering
        \includegraphics[width=\textwidth]{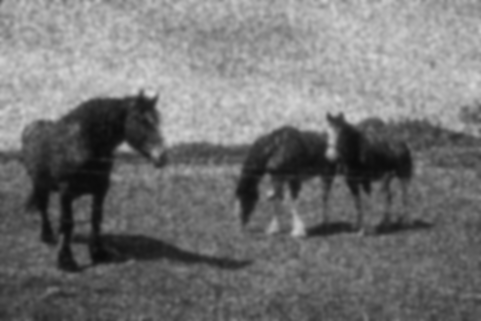}
        \caption{SHAN\cite{shan2019smooth}}
    \end{subfigure}
    \begin{subfigure}[b]{0.24\textwidth}
        \centering
        \includegraphics[width=\textwidth]{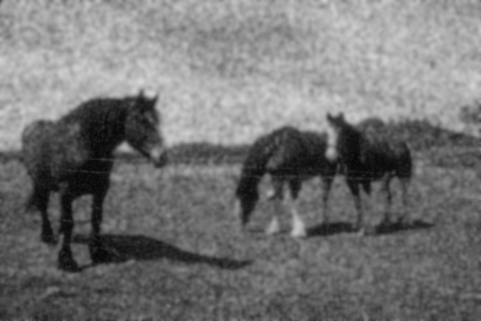}
        \caption{TDM\cite{Majee}}
    \end{subfigure}
    \begin{subfigure}[b]{0.24\textwidth}
        \centering
        \includegraphics[width=\textwidth]{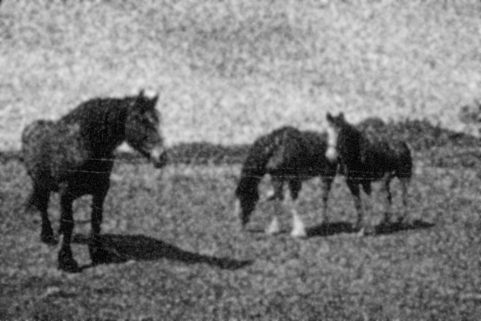}
        \caption{Proposed}
    \end{subfigure}

    \begin{subfigure}[b]{0.24\textwidth}
        \centering
        \includegraphics[width=\textwidth]{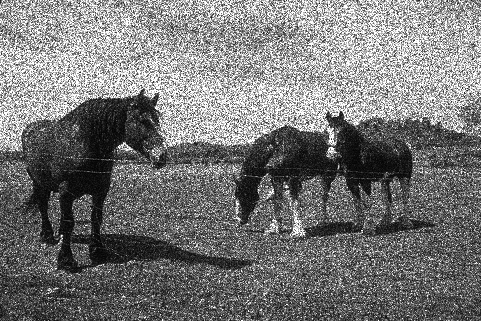}
        \caption{Look=5}
    \end{subfigure}
    \begin{subfigure}[b]{0.24\textwidth}
        \centering
        \includegraphics[width=\textwidth]{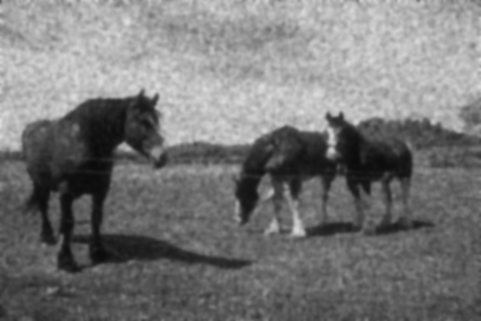}
        \caption{SHAN\cite{shan2019smooth}}
    \end{subfigure}
    \begin{subfigure}[b]{0.24\textwidth}
        \centering
        \includegraphics[width=\textwidth]{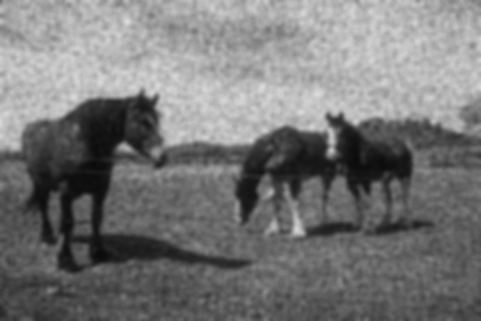}
        \caption{TDM\cite{Majee}}
    \end{subfigure}
    \begin{subfigure}[b]{0.24\textwidth}
        \centering
        \includegraphics[width=\textwidth]{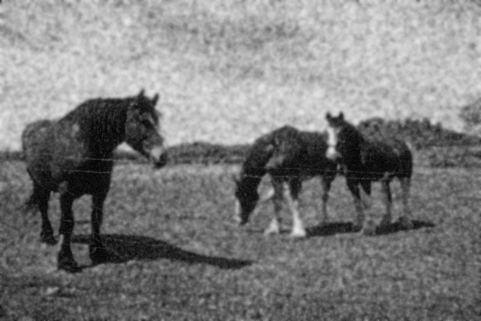}
        \caption{Proposed}
    \end{subfigure}

    \begin{subfigure}[b]{0.24\textwidth}
        \centering
        \includegraphics[width=\textwidth]{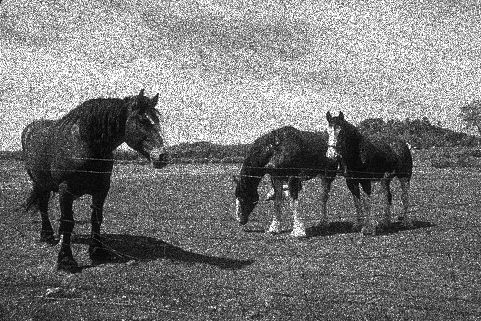}
        \caption{Look=10}
    \end{subfigure}
    \begin{subfigure}[b]{0.24\textwidth}
        \centering
        \includegraphics[width=\textwidth]{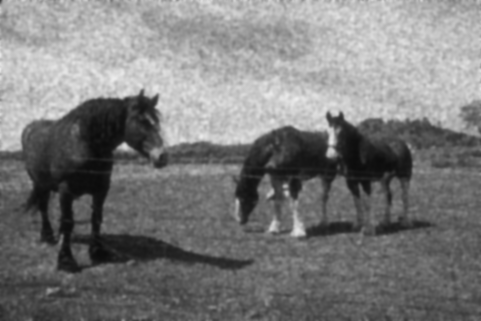}
        \caption{SHAN\cite{shan2019smooth}}
    \end{subfigure}
    \begin{subfigure}[b]{0.24\textwidth}
        \centering
        \includegraphics[width=\textwidth]{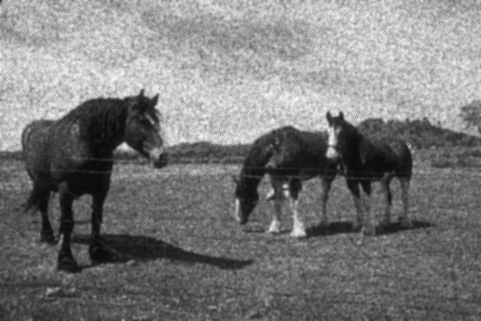}
        \caption{TDM\cite{Majee}}
    \end{subfigure}
    \begin{subfigure}[b]{0.24\textwidth}
        \centering
        \includegraphics[width=\textwidth]{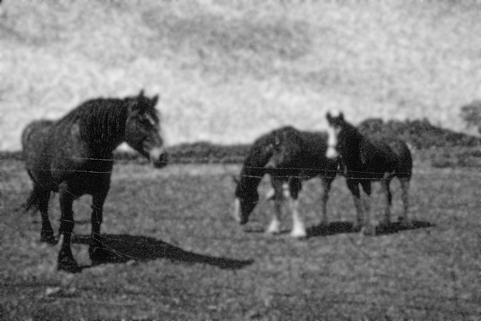}
        \caption{Proposed}
    \end{subfigure}

   \caption{
    The first column contains noisy pepper images with noise levels Look = 1, 3, 5, 10. 
    Subsequent columns: Restored test037 image using different models (SHAN\cite{shan2019smooth}, TDM\cite{Majee}, and Proposed Model).}
    \label{fig:noise_results2}
\end{figure}

A thorough investigation consisting of both visual and quantitative experiments was carried out to rigorously evaluate the despeckling capabilities of the proposed fourth-order telegraph diffusion model. The evaluation employed a diverse set of benchmark images, both grayscale and color modalities, corrupted by multiplicative speckle noise at different noise levels ($L = 1, 3, 5, 10$). Figures~\ref{fig:gamma_noise_results1}~\ref{fig:gamma_noise_results2}~\ref{fig:gamma_noise_results3}~\ref{fig:gamma_noise_results4}~\ref{fig:noise_results1}~\ref{fig:noise_results2} present the denoising results across grayscale images such as Boat, Texture, and a typical sample from the BSD68 dataset, as well as widely used color images Baboon and Peppers. Across all these examples, qualitative assessment reveals that the proposed method attains superior image restoration quality compared to second-order PDE-based despeckling methods. Further Speckle Index measurements in Table~\ref{tab:si_sar_imagesa} for SAR imagery experiments, where the notably lower speckle index values achieved by the proposed method confirm its superior despeckling efficiency.

\begin{table}[H]
    \centering
    \caption{Comparison of MSSIM and PSNR values for Baboon and Peppers images using Shan\cite{shan2019smooth}, TDE\cite{Majee}, and the Proposed model at different noise levels.}
    \label{tab:comparison5}
    \begin{tabular}{|c|c|cc|cc|cc|}
        \hline
        \multirow{2}{*}{\textbf{Image}} & \multirow{2}{*}{\textbf{Look}} 
        & \multicolumn{2}{c|}{\textbf{Shan\cite{shan2019smooth}}} 
        & \multicolumn{2}{c|}{\textbf{TDE\cite{Majee}}} 
        & \multicolumn{2}{c|}{\textbf{Proposed}} \\ 
        \cline{3-8}
        & & MSSIM & PSNR & MSSIM & PSNR & MSSIM & PSNR \\ 
        \hline
        \multirow{4}{*}{\textit{Baboon}}  
        & 1  & 0.5287 & 16.69 & 0.5386 & 16.71 & \textbf{0.5682} & \textbf{17.91} \\ 
        & 3  & 0.6221 & 19.07 & 0.6470 & 19.15 & \textbf{0.6833} & \textbf{19.51} \\ 
        & 5  & 0.6572 & 19.80 & 0.6722 & 19.89 & \textbf{0.7366} & \textbf{20.32} \\ 
        & 10 & 0.7438 & 21.11 & 0.7540 & 21.13 & \textbf{0.7998} & \textbf{21.60} \\ 
        \hline
        \multirow{4}{*}{\textit{Peppers}}  
        & 1  & 0.8155 & 17.55 & 0.8188 & 17.55 & \textbf{0.8515} & \textbf{20.47} \\ 
        & 3  & 0.9185 & 22.51 & 0.9187 & 22.51 & \textbf{0.9218} & \textbf{24.07} \\ 
        & 5  & 0.9401 & 24.52 & 0.9407 & 24.57 & \textbf{0.9408} & \textbf{25.51} \\ 
        & 10 & 0.9563 & 26.36 & 0.9577 & 26.56 & \textbf{0.9578} & \textbf{27.15} \\ 
        \hline
    \end{tabular}
\end{table}
\begin{table}[H]
\centering
\caption{Comparison of MSSIM and PSNR values for boat and texture images using Shan\cite{shan2019smooth}, TDE\cite{Majee}, and the Proposed model at different noise levels (Look values).}
\label{tab:comparison3}
\begin{tabular}{|l|l|ll|ll|ll|}
\hline
\multicolumn{1}{|c|}{images} & \multicolumn{1}{c|}{look}                                   & \multicolumn{2}{c|}{shan\cite{shan2019smooth}}                                                                                                                                                               & \multicolumn{2}{c|}{TDM\cite{Majee}}                                                                                                                                                                & \multicolumn{2}{c|}{Proposed}                                                                                                                                                              \\ \hline
                             &                                                             & \multicolumn{1}{l|}{MSSIM}                                                                              & PSNR                                                                          & \multicolumn{1}{l|}{MSSIM}                                                                              & PSNR                                                                          & \multicolumn{1}{l|}{\textbf{MSSIM}}                                                                       & \textbf{PSNR}                                                               \\ \hline
boat                         & \begin{tabular}[c]{@{}l@{}}1\\ 3\\ 5\\ 10\end{tabular} & \multicolumn{1}{l|}{\begin{tabular}[c]{@{}l@{}}0.4000\\ 0.4614\\ 0.4951\\ 0.5434\end{tabular}} & \begin{tabular}[c]{@{}l@{}}17.06\\ 22.29\\ 24.12\\ 25.89\end{tabular} & \multicolumn{1}{l|}{\begin{tabular}[c]{@{}l@{}}0.4116\\ 0.4803\\ 0.5208\\ 0.5702\end{tabular}} & \begin{tabular}[c]{@{}l@{}}17.06\\ 22.30\\ 24.17\\ 25.91\end{tabular} & \multicolumn{1}{l|}{\textbf{\begin{tabular}[c]{@{}l@{}}0.4524\\ 0.5726\\ 0.5992\\ 0.6109\end{tabular}}} & \textbf{\begin{tabular}[c]{@{}l@{}}20.86\\ 23.79\\ 24.65\\ 25.94\end{tabular}} \\ \hline

texture                      & \begin{tabular}[c]{@{}l@{}}1\\ 3\\ 5\\ 10\end{tabular} & \multicolumn{1}{l|}{\begin{tabular}[c]{@{}l@{}}0.5427\\ 0.5585\\ 0.5724\\ 0.5862\end{tabular}} & \begin{tabular}[c]{@{}l@{}}14.71\\ 19.28\\ 21.47\\ 24.07\end{tabular} & \multicolumn{1}{l|}{\begin{tabular}[c]{@{}l@{}}0.5453\\ 0.5729\\ 0.5874\\ 0.6032\end{tabular}} & \begin{tabular}[c]{@{}l@{}}14.71\\ 19.28\\ 21.47\\ 24.11\end{tabular} & \multicolumn{1}{l|}{\textbf{\begin{tabular}[c]{@{}l@{}}0.5463\\ 0.6092\\ 0.6454\\ 0.7730\end{tabular}}} & \textbf{\begin{tabular}[c]{@{}l@{}}17.20\\ 23.79\\ 24.65\\ 25.64\end{tabular}} \\ \hline
\end{tabular}
\end{table}
\begin{figure}[H]
    \centering
    \begin{minipage}{0.45\textwidth}
        \centering
        \includegraphics[width=\linewidth]{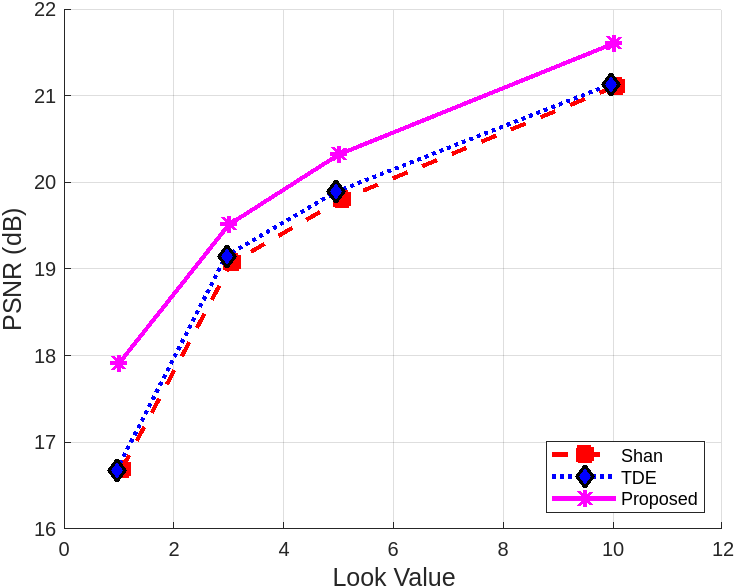}
        
    \end{minipage}
    \hfill
    \begin{minipage}{0.45\textwidth}
        \centering
        \includegraphics[width=\linewidth]{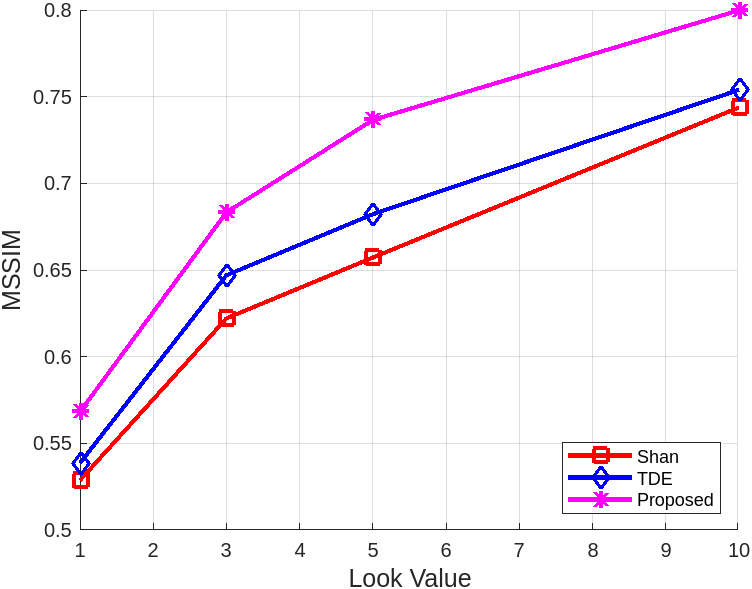}
       
    \end{minipage}
    
    \caption{Camparsion of PSNR and MSSIM  Proposed Model with State-of-art Models for the Color baboon image}
    \label{fig:psnr_mssim}
\end{figure}

\begin{figure}[H]
    \centering
    \begin{subfigure}[b]{0.25\textwidth}
        \centering
        \includegraphics[width=\textwidth, keepaspectratio]{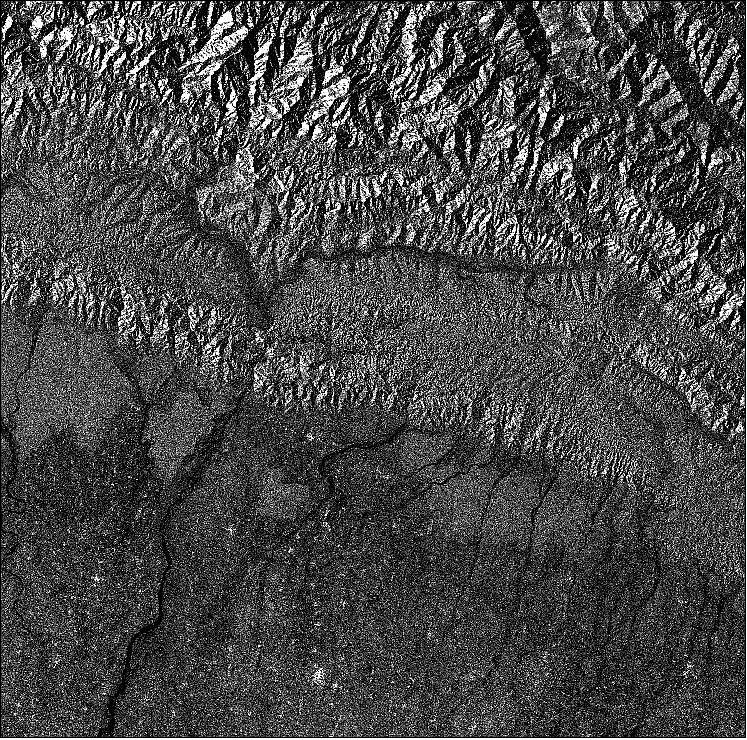}
        \caption*{Image 1}
    \end{subfigure}
    \begin{subfigure}[b]{0.25\textwidth}
        \centering
        \includegraphics[width=\textwidth, keepaspectratio]{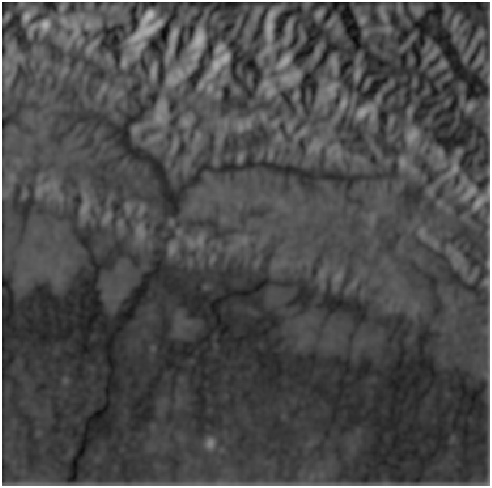}
        \caption*{TDM \cite{Majee}}
    \end{subfigure}
    \begin{subfigure}[b]{0.25\textwidth}
        \centering
        \includegraphics[width=\textwidth, keepaspectratio]{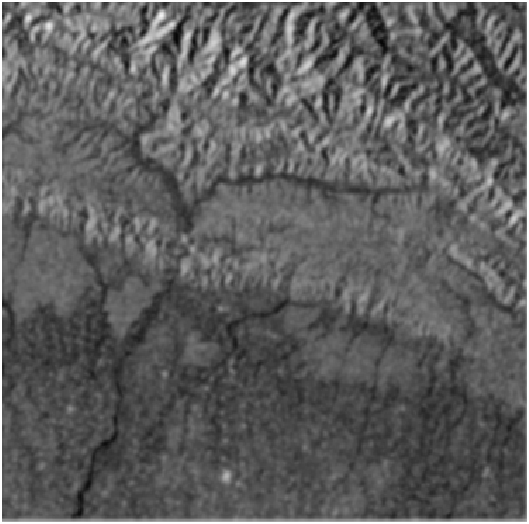}
        \caption*{Proposed }
    \end{subfigure}

    \begin{subfigure}[b]{0.25\textwidth}
        \centering
        \includegraphics[width=\textwidth, keepaspectratio]{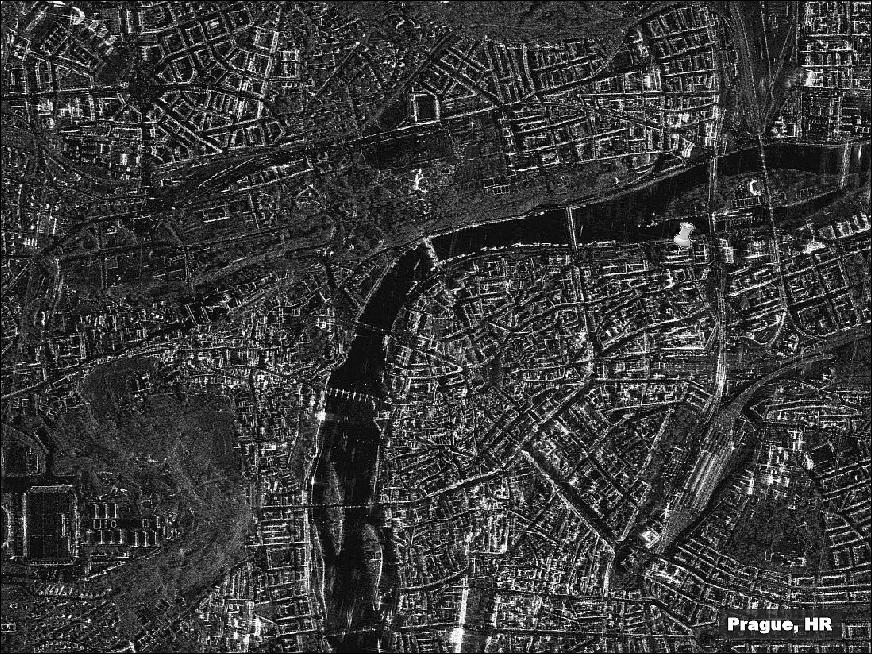}
        \caption*{Image 2}
    \end{subfigure}
    \begin{subfigure}[b]{0.25\textwidth}
        \centering
        \includegraphics[width=\textwidth, keepaspectratio]{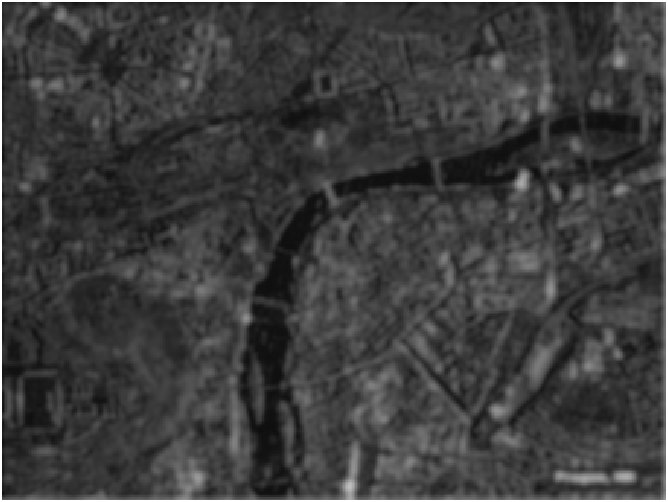}
        \caption*{TDM\cite{Majee}}
    \end{subfigure}
    \begin{subfigure}[b]{0.25\textwidth}
        \centering
        \includegraphics[width=\textwidth, keepaspectratio]{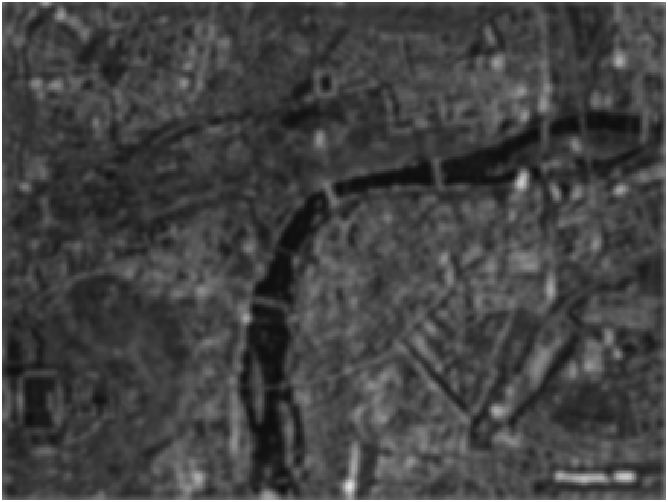}
        \caption*{Proposed}
    \end{subfigure}

    \caption{Comparison of SAR Image Restoration. Each row presents    (1) Noisy Image (2) TDM\cite{Majee} (3) Proposed.}
    \label{fig:sar_results}
\end{figure}

\begin{figure}[H]
    \centering
    
    \begin{subfigure}[b]{0.23\textwidth}
        \centering
        \includegraphics[width=\linewidth]{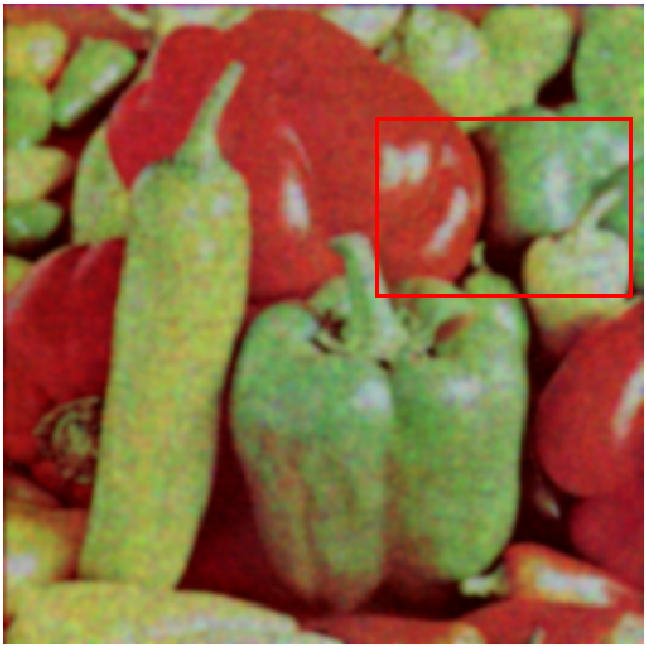}
        \caption{Shan\cite{shan2019smooth} }
    \end{subfigure}
    \begin{subfigure}[b]{0.23\textwidth}
        \centering
        \includegraphics[width=\linewidth]{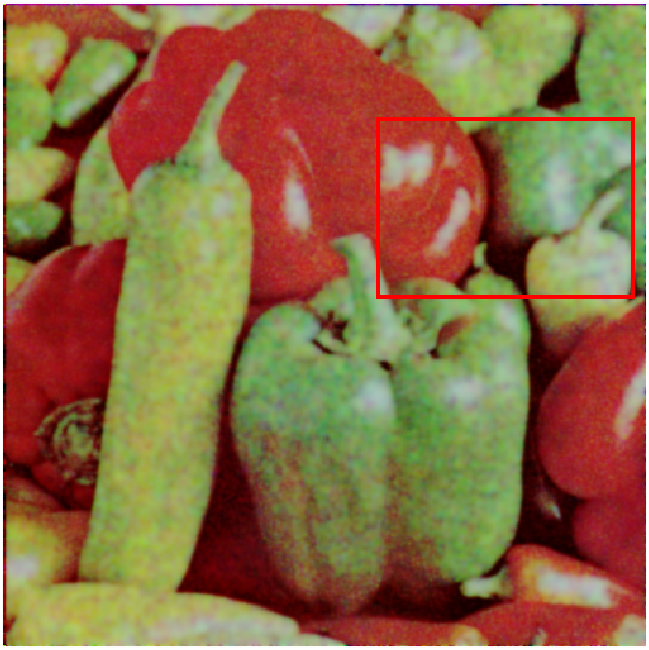}
        \caption{TDM\cite{Majee}}
    \end{subfigure}
    \begin{subfigure}[b]{0.23\textwidth}
        \centering
        \includegraphics[width=\linewidth]{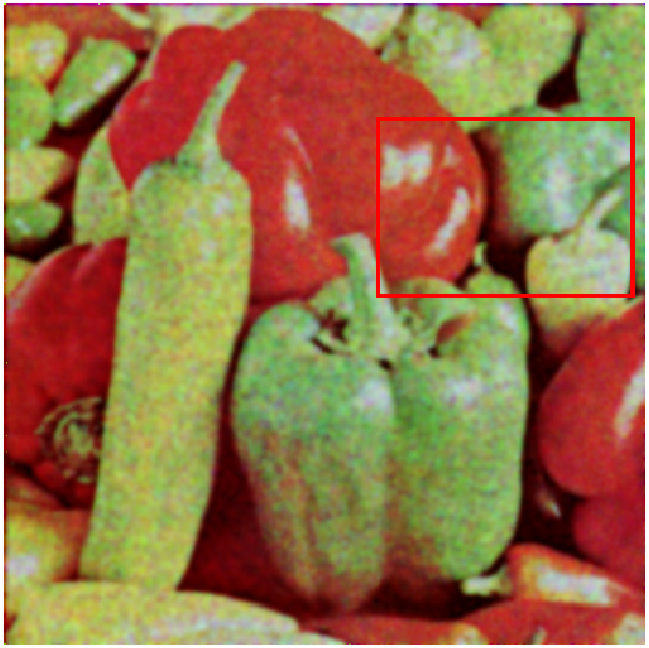}
        \caption{Proposed}
    \end{subfigure}

    \vspace{0.5cm} 

    \begin{subfigure}[b]{0.23\textwidth}
        \centering
        \includegraphics[width=\linewidth]{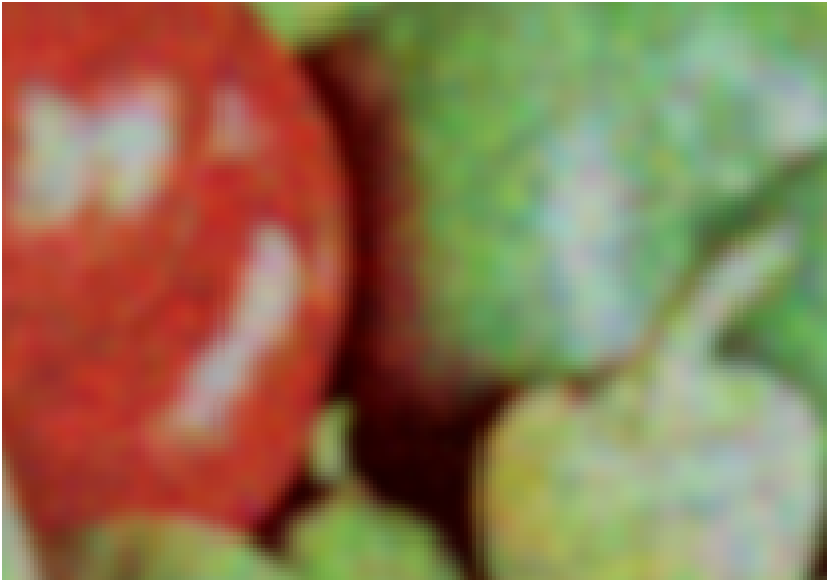}
        \caption{Shan\cite{shan2019smooth}}
    \end{subfigure}
    \begin{subfigure}[b]{0.23\textwidth}
        \centering
        \includegraphics[width=\linewidth]{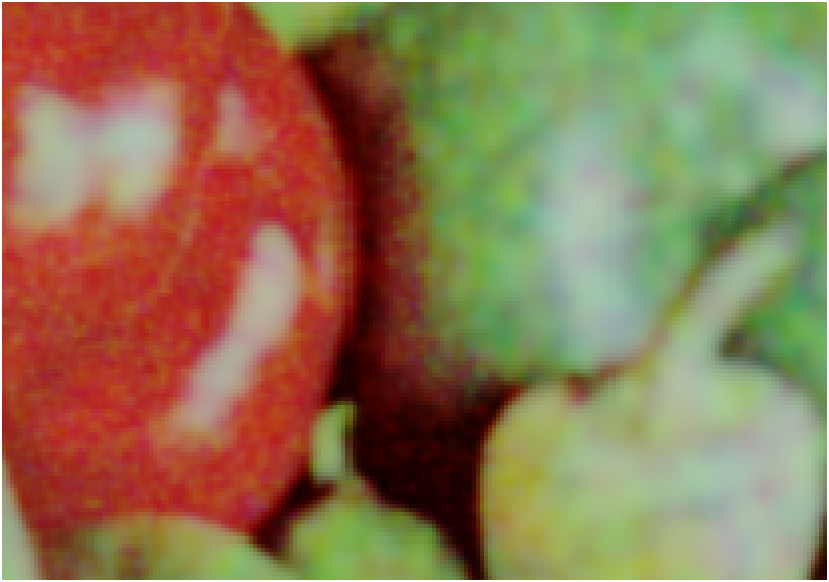}
        \caption{TDM\cite{Majee} }
    \end{subfigure}
    \begin{subfigure}[b]{0.23\textwidth}
        \centering
        \includegraphics[width=\linewidth]{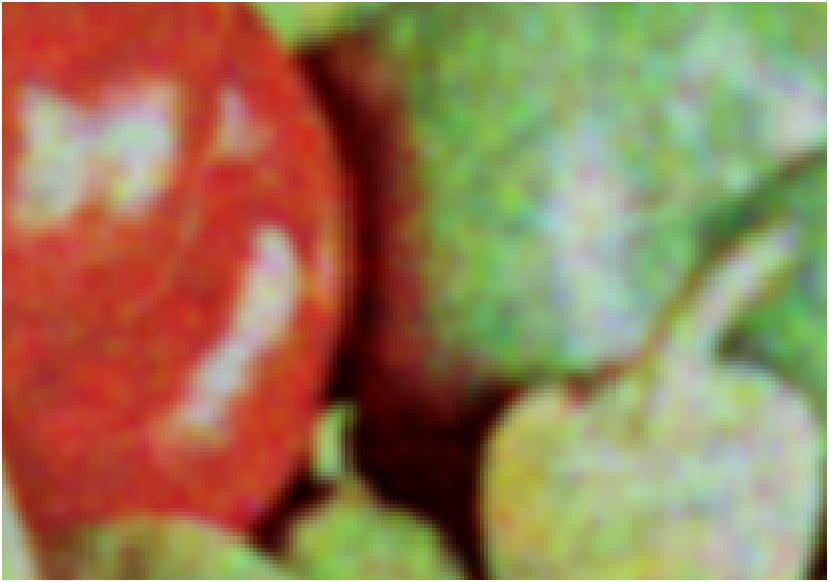}
        \caption{Proposed}
    \end{subfigure}

    \caption{Comparison of Denoising Performance: The top row displays the full Peppers images ($L=3$) with red boxes marking key despeckled areas, while the bottom row shows zoomed ROIs for a side-by-side evaluation of each model's noise removal effectiveness.}
    \label{fig:denoising_comparison}
    \end{figure}

Figure~\ref{fig:denoising_comparison} provides a zoomed-in visual comparison on a region of interest within the Peppers color image at $L=3$. The visual details reveal that the proposed model delivers markedly cleaner restorations, with preservation of edges and more accurate color fidelity relative to existing models, reinforcing its practical utility in real applications. Figure~\ref{fig:psnr_mssim} offers a comparative overview of PSNR and MSSIM across various methods and noise levels, underscoring the consistent dominance of the fourth-order model, especially in difficult scenarios such as the Baboon image.
 The results are listed in Tables~\ref{tab:comparison3} and ~\ref{tab:comparison4} for grayscale images, and in Table ~\ref{tab:comparison5} for color images. These numerical results consistently demonstrate that the fourth-order telegraph diffusion model significantly outperforms benchmark models such as the Shan and TDM methods. The improvements in PSNR and MSSIM values indicate the model’s enhanced ability to suppress noise while preserving essential image structures.
For fairness, all models parameter optimization tailored to grayscale and color image scenarios; optimal parameters utilized in the comparisons are summarized in Tables~\ref{tab:optimum_parameters1} . The robustness of the proposed model is particularly evident in its application to Synthetic Aperture Radar (SAR) images (Figure~\ref{fig:sar_results}), where it effectively reduces challenging speckle noise while preserving crucial geometric shapes and textural patterns that often degrade under conventional techniques.

\begin{table}[H]
\centering
\caption{Comparison of PSNR and MSSIM values for different models on BSD68 images under various Look levels.}
\label{tab:comparison4}

\setlength{\tabcolsep}{4pt}
\begin{tabular}{|l|l|c|c|c|c|c|c|}
\hline
\textbf{Image} & \textbf{Look} & \multicolumn{2}{c|}{\textbf{SHAN\cite{shan2019smooth}}} & \multicolumn{2}{c|}{\textbf{TDM\cite{Majee}}} & \multicolumn{2}{c|}{\textbf{Proposed}} \\
\cline{3-8}
              &               & MSSIM & PSNR & MSSIM & PSNR & MSSIM & PSNR \\ \hline

\multirow{4}{*}{test\_19} & 1  & 0.3129 & 16.75 & 0.3204 & 16.80 & \textbf{0.3736} & \textbf{18.20} \\
                         & 3  & 0.3910 & 20.20 & 0.4136 & 20.23 & \textbf{0.4964} & \textbf{21.04} \\
                         & 5  & 0.4247 & 21.57 & 0.4340 & 21.87 & \textbf{0.5471} & \textbf{22.21} \\
                         & 10 & 0.4636 & 23.09 & 0.4748 & 23.11 & \textbf{0.7141} & \textbf{23.94} \\ \hline
\multirow{4}{*}{test\_37} & 1  & 0.3763 & 16.21 & 0.3785 & 16.23 & \textbf{0.3894 }& \textbf{18.01} \\
                         & 3  & 0.3728 & 20.81 & 0.4295 & 20.89 & \textbf{0.5126 }& \textbf{21.17 }\\
                         & 5  & 0.3889 & 22.60 & 0.3984 & 22.69 & \textbf{0.5763 }& \textbf{23.31}\\
                         & 10 & 0.3633 & 23.07 & 0.3795 & 23.32 & \textbf{0.6560} & \textbf{23.91 }\\ \hline
\end{tabular}
\end{table}

\begin{table}[H]
\centering
\caption{S.I. for the Synthetic Aaprature Resonance Imags}
\begin{tabular}{cccc}
\hline
\textbf{Image} & \textbf{Noisy} & \textbf{Parameters} & \textbf{Proposed} \\ \hline
               &                & \(\boldsymbol{\gamma}  \quad\boldsymbol{\alpha} \quad \quad\boldsymbol{\lambda}\quad\quad \textbf{K}\) &                \\ \hline
Image\_1       & 1.02           & \(5 \quad 0.1 \quad 0.007 \quad 4\) & 0.2911           \\
Image\_2       & 0.92           & \(2 \quad 0.1 \quad 0.001 \quad 4\) & 0.3275           \\ \hline
\end{tabular}
\label{tab:si_sar_imagesa}
\end{table}

\begin{table}[H]
\centering
\caption{Optimum parameter values for different models on boat, texture, baboon, and pepper images}
\label{tab:optimum_parameters1}
\begin{tabular}{|l|c|c|c|c|c|c|c|c|c|c|}
\hline
\textbf{Image} & \textbf{Look} 
& \multicolumn{2}{c|}{\textbf{SHAN\cite{shan2019smooth}}} 
& \multicolumn{3}{c|}{\textbf{TDM\cite{Majee}}} 
& \multicolumn{4}{c|}{\textbf{Proposed}} \\
\hline
& \textbf{L} 
& $\alpha$ & $\beta$ 
& $\gamma$ & $\alpha$ & $k$ 
& $\gamma$ & $\alpha$ & $k$ & $\lambda$ \\
\hline
Boat 
& \begin{tabular}[c]{@{}l@{}}1\\3\\5\\10\end{tabular} 
& \begin{tabular}[c]{@{}c@{}}1\\1.1\\1.1\\1.2\end{tabular} 
& \begin{tabular}[c]{@{}c@{}}1\\1\\1\\1\end{tabular} 
& \begin{tabular}[c]{@{}c@{}}5\\5\\5\\5\end{tabular} 
& \begin{tabular}[c]{@{}c@{}}2\\2.5\\2.5\\2\end{tabular} 
& \begin{tabular}[c]{@{}c@{}}2\\2\\2\\2\end{tabular} 
& \begin{tabular}[c]{@{}c@{}}4\\4\\4\\4\end{tabular} 
& \begin{tabular}[c]{@{}c@{}}1\\1\\1\\3\end{tabular} 
& \begin{tabular}[c]{@{}c@{}}2\\2\\2\\2\end{tabular} 
& \begin{tabular}[c]{@{}c@{}}0.03\\0.03\\0.03\\0.03\end{tabular} \\
\hline
Texture 
& \begin{tabular}[c]{@{}l@{}}1\\3\\5\\10\end{tabular} 
& \begin{tabular}[c]{@{}c@{}}0.01\\0.1\\0.1\\0.4\end{tabular} 
& \begin{tabular}[c]{@{}c@{}}0.01\\0.01\\0.01\\0.01\end{tabular} 
& \begin{tabular}[c]{@{}c@{}}4\\4\\2\\2\end{tabular} 
& \begin{tabular}[c]{@{}c@{}}0.1\\1\\1\\1\end{tabular} 
& \begin{tabular}[c]{@{}c@{}}2\\2\\2\\2\end{tabular} 
& \begin{tabular}[c]{@{}c@{}}4\\4\\4\\4\end{tabular} 
& \begin{tabular}[c]{@{}c@{}}0.4\\1\\0.5\\0.5\end{tabular} 
& \begin{tabular}[c]{@{}c@{}}2\\2\\2\\2\end{tabular} 
& \begin{tabular}[c]{@{}c@{}}0.03\\0.03\\0.03\\0.05\end{tabular} \\
\hline
baboon & \begin{tabular}[c]{@{}l@{}}1\\ 3\\ 5\\ 10\end{tabular} & \multicolumn{1}{l|}{\begin{tabular}[c]{@{}l@{}}1\\ 1\\ 1.1\\ 1.2\end{tabular}} & \begin{tabular}[c]{@{}l@{}}1\\ 1\\ 1\\ 1\end{tabular} & \multicolumn{1}{l|}{\begin{tabular}[c]{@{}l@{}}2.5\\ 3\\ 3.1\\ 3.1\end{tabular}}     & \multicolumn{1}{l|}{\begin{tabular}[c]{@{}l@{}}2\\ 2\\ 2\\ 2\end{tabular}} & \begin{tabular}[c]{@{}l@{}}9\\ 5\\ 5\\ 5\end{tabular} & \multicolumn{1}{l|}{\begin{tabular}[c]{@{}l@{}}1\\ 1.1\\ 1.2\\ 1.2\end{tabular}} & \multicolumn{1}{l|}{\begin{tabular}[c]{@{}l@{}}2\\ 2\\ 2\\ 2\end{tabular}} & \multicolumn{1}{l|}{\begin{tabular}[c]{@{}l@{}}5\\ 5\\ 5\\ 5\end{tabular}} & \begin{tabular}[c]{@{}l@{}}0.1\\ 0.1\\ 0.1\\ 0.1\end{tabular}  \\ \hline
pepper & \begin{tabular}[c]{@{}l@{}}1\\ 3\\ 5\\ 10\end{tabular} & \multicolumn{1}{l|}{\begin{tabular}[c]{@{}l@{}}1\\ 1\\ 1.1\\ 1.2\end{tabular}} & \begin{tabular}[c]{@{}l@{}}1\\ 1\\ 1\\ 1\end{tabular} & \multicolumn{1}{l|}{\begin{tabular}[c]{@{}l@{}}2\\ 2.1\\ 2.5\\ 2.5\end{tabular}} & \multicolumn{1}{l|}{\begin{tabular}[c]{@{}l@{}}2\\ 2\\ 2\\ 2\end{tabular}} & \begin{tabular}[c]{@{}l@{}}5\\ 9\\ 9\\ 9\end{tabular} & \multicolumn{1}{l|}{\begin{tabular}[c]{@{}l@{}}1\\ 1\\ 1\\ 2\end{tabular}}       & \multicolumn{1}{l|}{\begin{tabular}[c]{@{}l@{}}2\\ 2\\ 2\\ 2\end{tabular}} & \multicolumn{1}{l|}{\begin{tabular}[c]{@{}l@{}}5\\ 5\\ 5\\ 5\end{tabular}} & \begin{tabular}[c]{@{}l@{}}0.1\\ 0.1\\ 0.1\\ 0.08\end{tabular} \\ \hline
\end{tabular}
\end{table}
\section{Conclusion}\label{sec:Conclusion}
This work introduces a novel fourth-order telegraph diffusion model designed to address the challenges of multiplicative speckle noise in image denoising. The proposed model effectively balances noise reduction and feature preservation, mitigating common issues such as staircase artifacts that are prevalent in second-order models. Through extensive numerical simulations, the model demonstrates superior performance in maintaining intricate details such as contours and textures while efficiently suppressing noise. The results highlight the model's to existing state-of-the-art second-order, particularly in applications involving noisy SAR images. Also, we prove the well-posedness of the proposed model. Overall, the fourth-order PDE model represents a significant advancement in image despeckling, offering a robust solution for handling noisy images in various real-world applications.

\vspace{11pt}
\noindent\textbf{Author Declaration} \\
The authors declare that there are no conflicts of interest.

\vspace{11pt}
\noindent\textbf{Data Availability} \\
The data that support the findings of this study are available from the corresponding
author upon reasonable request.

\end{document}